%% file: main.tex
\documentclass{article}

\PassOptionsToPackage{numbers, compress}{natbib}

\usepackage[preprint]{neurips_2026}

\usepackage[utf8]{inputenc} 
\usepackage[T1]{fontenc}    
\usepackage{hyperref}       
\usepackage{url}            
\usepackage{booktabs}       
\usepackage{amsfonts}       
\usepackage{nicefrac}       
\usepackage{microtype}      
\usepackage{xcolor}         
\usepackage{amsmath}
\usepackage{graphicx}
\usepackage{subcaption}
\usepackage{graphicx}
\usepackage{siunitx}
\usepackage{tabularx}
\usepackage{placeins}

\input{preamable}
\title{TiCo: Time-Controllable Spoken Dialogue Model}

\author{%
\begin{tabular}{c}
Kai-Wei Chang$^{\spadesuit}$\thanks{Equal contribution} \hspace{1.2em}
Wei-Chih Chen$^{\diamondsuit}$\footnotemark[1] \hspace{1.2em}
En-Pei Hu$^{\diamondsuit}$ \hspace{1.2em}
Hung-yi Lee$^{\diamondsuit \clubsuit}$ \hspace{1.2em}
James Glass$^{\spadesuit}$ \\
[0.5ex]\\
{\normalfont
$^{\spadesuit}$ MIT \hspace{1.5em}
$^{\diamondsuit}$ NTU \hspace{1.5em}
$^{\clubsuit}$ NTU AI-CoRE \hspace{1.5em}
}
\\
\\
{\normalfont
kwchang@mit.edu
}
\end{tabular}
}

\begin{document}

\maketitle

\begin{abstract}
We introduce \textbf{TiCo}, a time-controllable spoken dialogue model (SDM) that follows time-constrained instructions (e.g., ``Please generate a response lasting about 15 seconds'') and generates spoken responses with controllable duration.
This capability is valuable for real-world spoken language systems such as voice assistants and interactive agents, where controlling response duration can improve interaction quality.
However, despite their strong ability to generate natural spoken responses, existing models lack time awareness and struggle to follow duration-related instructions. 
To systematically evaluate this, we introduce TiCo-Bench, the first benchmark for time-controllable instruction following in SDMs, on which existing open-source and commercial models frequently fail to satisfy explicit time constraints.
TiCo addresses this limitation by enabling an SDM to estimate elapsed speaking time during generation through Spoken Time Markers (STM) (e.g., \texttt{<10.6 seconds>}). 
These markers help the model maintain awareness of time and adjust the remaining content to meet the target duration.
TiCo is post-trained efficiently without question-answer paired data, relying on self-generation and reinforcement learning with verifiable reward.
Experimental results show that TiCo reduces duration error by 2.7$\times$ over its backbone and 1.6$\times$ over the strongest baseline, while preserving response quality.
\end{abstract}

\input{Sections/Introduction}
\input{Sections/Related_works}
\input{Sections/Method}
\input{Sections/Experimental-Setup}
\input{Sections/Results}
\input{Sections/Conclusion}

\bibliographystyle{unsrtnat}
\bibliography{references}

\newpage
\appendix
\input{Sections/Appendix}


\end{document}

%% file: preamable.tex
\usepackage[table]{xcolor}
\usepackage{array}
\usepackage[most]{tcolorbox}

\definecolor{commercialcolor}{RGB}{255,235,210}
\definecolor{opensourcecolor}{RGB}{220,235,255}
\definecolor{cascadedcolor}{RGB}{225,245,225}
\definecolor{proposedcolor}{RGB}{235,225,255}

\newcommand{\modeltag}[2]{%
\tcbox[
    enhanced,
    colback=#1,
    colframe=#1,
    boxrule=0pt,
    arc=2.5pt,
    left=4pt,
    right=4pt,
    top=2pt,
    bottom=2pt,
    boxsep=0pt,
    width=\linewidth,
    nobeforeafter,
    valign=center
]{#2}%
}

\newcommand{\commercialbox}[1]{\modeltag{commercialcolor}{#1}}
\newcommand{\opensourcebox}[1]{\modeltag{opensourcecolor}{#1}}
\newcommand{\cascadedbox}[1]{\modeltag{cascadedcolor}{#1}}
\newcommand{\proposedbox}[1]{\modeltag{proposedcolor}{#1}}

\newcommand{\legenddot}[1]{%
\raisebox{0.2ex}{\colorbox{#1}{\rule{0pt}{0.9ex}\hspace{0.9em}}}%
}

%% file: Sections/Introduction.tex
\input{Figures/TiCo-overview.drawio}
\section{Introduction}

\emph{``Time is money,''} as famously stated by Benjamin Franklin, highlights the fundamental value of time in human life.
In human–computer interaction, time is a critical resource that directly impacts usability, deployment cost, and safety-critical decision making.
This is especially true for \emph{Spoken Dialogue Models} (SDMs)~\cite{aroralandscape, cui2025recent, ji2024wavchat}, which are gaining increasing attention in real-world applications, such as personal assistants, wearable devices, and healthcare systems~\cite{de2020intelligent, adams2025generative}.
In these scenarios, a response must not only be accurate and natural, but often also strictly bounded in duration. 
For example, a voice assistant may be required to provide a traffic update while driving; a wearable device may require concise spoken feedback due to battery or bandwidth constraints. Similarly, in medical or emergency scenarios, a voice assistant may need to deliver brief yet informative instructions under strict time pressure. 
In all of these cases, the ability to control response duration is a key requirement for practical deployment. 
Despite its importance, time controllability remains largely underexplored in SDMs.

In the domain of text Large Language Models (LLMs), prior studies have shown that models often struggle to follow explicit length-constraint instructions~\cite{zhang2025lifebench}. 
Moreover, LLM outputs often exhibit verbosity or length bias, a phenomenon associated with preference-based evaluation and alignment~\cite{dubois2024length, hu-etal-2025-explaining}.
This tendency weakens instruction-following capability and negatively affects user experience. 
While benchmarks, prompting and training strategies have begun to address length controllability in text LLMs~\cite{zhang2025lifebench, xie2025prompt, jie2023prompt, song2025hansel}, this research direction remains active and continues to attract attention due to its substantial practical importance.

However, controlling response duration in SDMs is considerably more challenging than controlling output length in text. 
In speech generation, word count is only a proxy for actual duration. 
A single word may contain different numbers of syllables, and speech duration is known to vary with phonetic composition, linguistic context, and prosodic structure~\cite{klatt1976linguistic}. 
Moreover, speaking rate may vary across speakers, speaking styles, and communicative conditions, which depends on both speaker and listener~\cite{lindblom1990explaining}. 
As a result, simply constraining the number of generated words does not guarantee accurate control over the final speech duration. 
This mismatch makes duration control a unique and more demanding problem in spoken dialogue systems.

Given the limited study of time controllability in SDMs, we first introduce \textbf{TiCo-Bench}, a benchmark designed to evaluate the time-controllable instruction-following capability of SDMs. Our evaluation reveals that existing SDMs struggle to reliably satisfy explicit time constraints.

To address this challenge, we introduce \textbf{TiCo} (Figure~\ref{fig:TiCo-overview}), a time-controllable SDM that can estimate and regulate generated speech duration in real time through \textbf{Spoken Time Markers}. TiCo is obtained by post-training an SDM to develop an internal mechanism for \emph{time awareness} during generation, enabling it to track temporal progress and adjust its responses accordingly.

Specifically, TiCo is trained with a two-stage procedure.
In the first stage, the model leverages \emph{self-generation} to construct supervision data for learning duration awareness, enabling it to associate intermediate generation states with temporal progress and estimate the elapsed speaking time.
In the second stage, \emph{Reinforcement Learning with Verifiable Rewards (RLVR)}~\cite{lambert2024tulu, shao2024deepseekmath} is applied, where rewards are automatically verified based on the Spoken Time Markers, to further shape the response distribution and improve compliance with duration-related instructions. This stage encourages the model to better satisfy target time constraints while preserving the response quality, including helpfulness and coherence.

Our contributions are summarized as follows:
\begin{itemize}
\item We propose \textbf{TiCo}, a time-controllable spoken dialogue model trained with a two-stage post-training procedure, enabling it to generate \textbf{Spoken Time Markers (STMs)} during inference and perform real-time control over response duration.
\item We introduce \textbf{TiCo-Bench}, the first benchmark designed to evaluate the time controllability of spoken dialogue models, measuring whether they can follow explicit duration-related instructions.
\item We conduct extensive experiments showing that TiCo significantly improves duration controllability while preserving response quality, and further demonstrate that the learned capability generalizes beyond the duration range seen during training.
\end{itemize}

%% file: Figures/TiCo-overview.drawio.tex
\begin{figure*}[h]
    \centering
    \includegraphics[width=\textwidth]{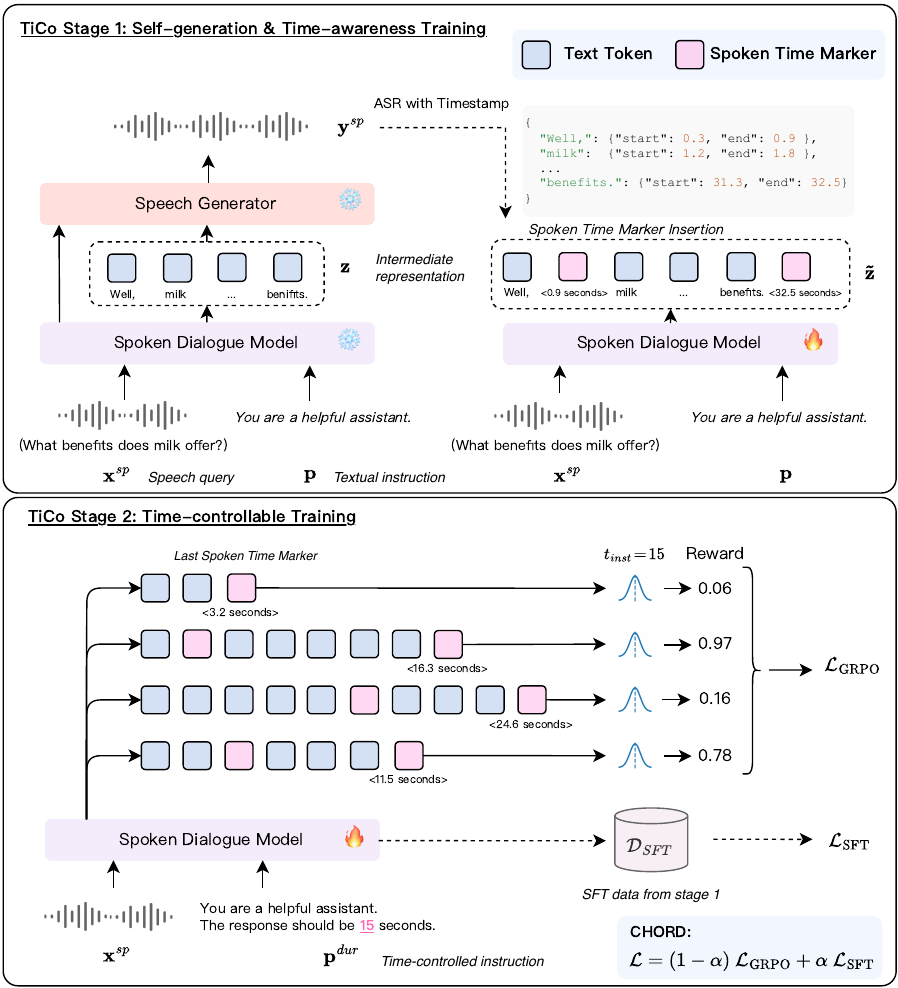}
\caption{Overview of \textbf{TiCo}, a two-stage framework for time-controllable speech generation. 
\textbf{Stage 1 (top):} The model leverages self-generation to produce responses annotated with \textbf{Spoken Time Markers (STMs)}, which serve as supervision for learning \emph{time awareness}, i.e., associating intermediate generation states with temporal progress and estimating elapsed speaking time. 
\textbf{Stage 2 (bottom):} The model is further optimized via RLVR, where rewards are derived from STMs, enabling the model to regulate response duration in real time.}
    \label{fig:TiCo-overview}
\end{figure*}

%% file: Sections/Related_works.tex
\section{Related Works}

\subsection{Spoken Dialogue Models}
Spoken Dialogue Models (SDMs)~\cite{aroralandscape, cui2025recent} aim to enable natural human-computer interaction by directly understanding and generating spoken conversations. 
Unlike traditional voice assistants that rely on cascaded ASR, text generation, and TTS modules, recent SDMs increasingly adopt end-to-end or tightly integrated modeling paradigms~\cite{glass1999challenges, ji2024wavchat}.

However, compared to text-based LLMs operating in a semantically rich textual space, speech is considered to be significantly more challenging to process due to the high variability and complexity of acoustic signals\footnote{This challenge has been largely explored in prior work such as the \emph{``Textless NLP'' paradigm}~\cite{shao2024deepseekmath, polyak2021speech, hassid2023textually, chang2022speechprompt, chang2024speechprompt, popuri2022enhanced}, where quantized speech representations are treated as ``pseudo text'' to improve training efficiency and efficacy.}. 
As a result, many recent SDMs introduce \emph{intermediate representations}, most commonly text, to support \emph{semantic planning} during generating the speech response. 
This includes reasoning to improve response quality~\cite{chiang2026stitch}, tool calling~\cite{arora2025stream} to leverage external modules, and more direct guidance over spoken content~\cite{xu2025qwen25, xu2025qwen3}. 
Specifically, the SDM first takes the input query (in either text or speech form) to generate an intermediate representation, which is then consumed by a speech generator to produce the final output speech representation (e.g., phonetic tokens and acoustic tokens~\cite{aroralandscape, guo2025recent, wu2024codec}), and subsequently synthesized into a waveform.
We provide a survey of representative SDMs and their intermediate representations in the Appendix~\ref{Appendix:Survey}.

Recently, several benchmarks have begun to evaluate SDMs beyond response quality, incorporating dimensions such as speaking style~\cite{yang2025paras2s}, interactivity~\cite{lin2025full_v1, lin2025full_v2}, controllability~\cite{yang2025paras2s} and time-awareness~\cite{chang2025game}. Despite the emergence of such benchmarks on controllability and time-awareness, to the best of our knowledge, TiCo is the first method that explicitly enables \emph{time-controllable} generation for SDMs through an efficient post-training approach.


It is worth noting that TiCo differs fundamentally from \emph{duration modeling} in TTS systems~\cite{renfastspeech, xie2025towards}.
While duration modeling in TTS primarily focuses on aligning text with synthesized speech, TiCo instead targets time-controllable spoken response generation. This setting requires spoken dialogue models (SDMs) to perform semantic planning and reasoning while dynamically adapting to time-related constraints during generation.
Moreover, TiCo is orthogonal to prior work on \emph{temporal understanding}~\cite{sridhar2025enhancing}, which aims to equip speech models with the ability to interpret temporal information in input audio (e.g., ``What is the time interval of the query `a dog barking' in the audio?'')~\cite{ wang2025listening}.
In contrast, TiCo focuses on time awareness in the \emph{generation process}, rather than temporal comprehension of the input.

\subsection{Length-Control Large Language Models}

Length control in text LLMs has been studied along three lines. 
Training-free or decoding-time methods enforce length via 
sampling~\cite{gu2025length}, zero-shot prompting~\cite{retkowski2025zero}, 
or EOS-token reweighting~\cite{belligoli2025controlling}. Instruction-tuning 
approaches inject length-tracking signals into generation, such as 
distance-to-target encodings~\cite{butcher2025precise}, latent tracking 
tokens~\cite{song2025hansel}, or explicit positional 
markers~\cite{wang2024positionid}. A third line uses RL or preference 
optimization to decouple length bias from response 
quality~\cite{liu2024length, li2025length} and to control reasoning length, 
either by enforcing concise steps~\cite{liu2026laconic} or by extending the 
trajectory for harder problems~\cite{liu2025prorl, aggarwal2025l1, 
jie2023prompt}. All these methods operate on text, where word count is a 
direct proxy for length. TiCo instead targets spoken dialogue, where word 
count is only a loose proxy for speech duration since the realized timing 
depends on paralinguistic factors beyond the textual output.

%% file: Sections/Method.tex
\section{TiCo}
\label{sec:tico-method}
A speech-to-speech Spoken Dialogue Model (SDM) can be viewed as a conditional generative model that produces a spoken response $\mathbf{y}^{\mathrm{sp}}$ given the user's input speech query $\mathbf{x}^{\mathrm{sp}}$
and a textual instruction $\mathbf{p}$ (e.g., a system prompt). 


Modern SDMs often introduce \emph{intermediate representations} $\mathbf{z}$ to bridge high-level semantic reasoning and low-level speech synthesis. 
Concretely, an \textbf{intermediate sequence generator} $p_{\theta}$ first generates an intermediate representation conditioned on the user input:
\begin{equation}
\mathbf{z} \sim p_{\theta}(\mathbf{z} \mid \mathbf{x}^{\mathrm{sp}}, \mathbf{p}).
\end{equation}

The final spoken response is then generated by a \textbf{speech generator} $q_{\phi}$:
\begin{equation}
\mathbf{y}^{\mathrm{sp}} \sim q_{\phi}(\mathbf{y}^{\mathrm{sp}} \mid \mathbf{z}, \mathbf{x}^{\mathrm{sp}}, \mathbf{p}).
\end{equation}
Different architectures impose different conditional independence assumptions on Eq.~(2). In cascaded systems, the speech synthesis module has no access to the original user speech or instruction, reducing the generation to $q_{\phi}(\mathbf{y}^{\mathrm{sp}} \mid \mathbf{z})$. In end-to-end models, the generation of $\mathbf{y}^{\mathrm{sp}}$ may additionally depend on $\mathbf{x}^{\mathrm{sp}}$ and $\mathbf{p}$ \footnote{For example, in Qwen-Omni's ``Thinker-Talker'' design~\cite{xu2025qwen25, xu2025qwen3}}.

\subsection{TiCo Stage1: Time-Awareness Training}
This stage (Figure~\ref{fig:TiCo-overview} (top)) trains the model to generate \emph{Spoken Time Markers} as part of the intermediate representation \(\mathbf{z}\), so that \(\mathbf{z}\) encodes not only semantic content but also its expected temporal alignment with the final spoken response \(\mathbf{y}^{\mathrm{sp}}\) under the conditioning context \((\mathbf{x}^{\mathrm{sp}}, \mathbf{p})\).
These markers are inserted into $\mathbf{z}$ through a self-generation process and used as prediction targets during training.

\noindent
\textbf{Spoken Time Marker.}
A Spoken Time Marker is a special token indicating the estimated cumulative speaking duration up to a given position in the intermediate representation. Conceptually, these markers serve as a discretized alignment signal between the intermediate semantic plan \(\mathbf{z}\) and the realized spoken response \(\mathbf{y}^{\mathrm{sp}}\) under the same conditioning context \((\mathbf{x}^{\mathrm{sp}}, \mathbf{p})\). 
Inspired by TimeMarker~\cite{chen2024timemarker}, we represent these markers in textual form, e.g., \texttt{<6.8 seconds>}.


Estimating duration at the intermediate level is non-trivial. A single word may correspond to multiple syllables, and its acoustic duration may vary depending on context and speaking rate. 
Explicit duration estimation is therefore required to bridge the gap between the intermediate representation and the final speech realization.

\noindent
\textbf{Training Data Construction.}
Let \(\mathcal{D}=\{(\mathbf{x}^{\mathrm{sp}}, \mathbf{p})\}\) denote a pool of input speech query--instruction pairs. In this stage, we construct time-aware training targets through \emph{self-generation} followed by ASR-based alignment. Specifically, given each input \((\mathbf{x}^{\mathrm{sp}}, \mathbf{p}) \in \mathcal{D}\), the model first freely generates an intermediate representation \(\mathbf{z}\) and its corresponding spoken response \(\mathbf{y}^{\mathrm{sp}}\). 

We then apply ASR-based alignment to estimate the temporal correspondence between \(\mathbf{z}\) and \(\mathbf{y}^{\mathrm{sp}}\). Based on the aligned timestamps, we define a sequence of Spoken Time Markers \(\mathbf{t} = [t_1, \dots, t_M]\), where each \(t_j\) denotes the estimated cumulative speaking duration at an aligned position in \(\mathbf{z}\). We interleave these markers with the intermediate tokens to obtain an augmented sequence:
\begin{equation}
\tilde{\mathbf{z}} = [\, z_1, \dots, z_i, t_j, \dots, z_N, t_M \,].
\end{equation}

As a result, the augmented sequence \(\tilde{\mathbf{z}}\) encodes not only semantic content, but also alignment-induced timing information that links \(\mathbf{z}\) to the final spoken response under the same input condition \((\mathbf{x}^{\mathrm{sp}}, \mathbf{p})\).

This process yields an aligned training set
$
\mathcal{D}_{\mathrm{SFT}}=\{(\mathbf{x}^{\mathrm{sp}}, \mathbf{p}, \tilde{\mathbf{z}})\}.
$
We model the augmented intermediate sequence autoregressively as
\begin{equation}
p_{\theta}(\tilde{\mathbf{z}} \mid \mathbf{x}^{\mathrm{sp}}, \mathbf{p})
=
\prod_{n=1}^{|\tilde{\mathbf{z}}|}
p_{\theta}(\tilde{\mathbf{z}}_{n} \mid \tilde{\mathbf{z}}_{<n}, \mathbf{x}^{\mathrm{sp}}, \mathbf{p}).
\end{equation}
We then optimize the standard supervised fine-tuning (SFT) objective:
\begin{equation}
\mathcal{L}_{\mathrm{SFT}}
=
-\mathbb{E}_{(\mathbf{x}^{\mathrm{sp}}, \mathbf{p}, \tilde{\mathbf{z}})\sim \mathcal{D}_{\mathrm{SFT}}}
\left[
\sum_{n=1}^{|\tilde{\mathbf{z}}|}
\log p_{\theta} \left(\tilde{\mathbf{z}}_{n} \mid \tilde{\mathbf{z}}_{<n}, \mathbf{x}^{\mathrm{sp}}, \mathbf{p}\right)
\right].
\end{equation}

It's worth noting that self-generation offers two advantages:
(1)~it removes the need for collecting paired question-answer supervision, and
(2)~the generated responses follow the model's own output distribution, which improves training stability~\cite{lu2025desta2}.

\subsection{TiCo Stage 2: Time-Controllable Training}

This stage (Figure~\ref{fig:TiCo-overview} (bottom)) further trains the model to follow time-constrained instructions. We augment the textual instruction \(\mathbf{p}\) with a duration constraint and denote the resulting instruction by \(\mathbf{p}^{\mathrm{dur}}\), where the target duration is denoted by \(t_{\mathrm{inst}}\). Since Spoken Time Markers reside in the intermediate representation, we apply reinforcement learning to the intermediate-sequence generator \(p_{\theta}(\tilde{\mathbf{z}} \mid \mathbf{x}^{\mathrm{sp}}, \mathbf{p}^{\mathrm{dur}})\).

Specifically, we adopt GRPO~\cite{shao2024deepseekmath} to optimize time controllability, and incorporate CHORD~\cite{zhang2026onpolicyrlmeetsoffpolicy} as a dynamically weighted auxiliary objective that integrates off-policy expert trajectories into the on-policy RL process. This regularization steers the policy toward the expert trajectories in the Stage-1-constructed dataset $\mathcal{D}_{\mathrm{SFT}}$ while preserving on-policy exploration. In practice, we find this mechanism crucial for stabilizing training, as GRPO alone frequently leads to reward hacking.

Given an input \((\mathbf{x}^{\mathrm{sp}}, \mathbf{p}^{\mathrm{dur}})\), we sample a group of \(G\) candidate augmented intermediate sequences from the old policy:
\begin{equation}
\tilde{\mathbf{z}}^{(g)} \sim p_{\theta_{\mathrm{old}}}(\cdot \mid \mathbf{x}^{\mathrm{sp}}, \mathbf{p}^{\mathrm{dur}}),
\qquad g=1,\dots,G.
\end{equation}

\textbf{Reward Design.}
The main reward measures the accuracy of the predicted total duration:
\begin{equation}
\mathcal{R}_{\text{main}}^{(g)}
= F \left(t_{\text{inst}} - t_{\text{last}}^{(g)}\right),
\end{equation}
where \(t_{\text{inst}}\) is the target duration specified in the instruction and \(t_{\text{last}}^{(g)}\) is the duration indicated by the final generated time marker in \(\tilde{\mathbf{z}}^{(g)}\). We instantiate \(F\) as a Gaussian function, i.e., \(F(\Delta t) = \exp\left(-(\Delta t)^2 / (2\sigma^2)\right)\), where \(\sigma\) controls the tolerance to duration errors.

We additionally introduce several auxiliary rewards to stabilize training and mitigate \emph{reward hacking}, including a \textbf{``presence reward''} that encourages the model to generate at least one time marker, a \textbf{``monotonicity reward''} that encourages time markers to increase monotonically, a \textbf{``repetition penalty''} that discourages repeatedly generating identical time markers, and a \textbf{``copy penalty''} that discourages trivial copying of the instructed duration. The detailed definitions of these auxiliary rewards are provided in Appendix~\ref{app:training_details} and ablation study is in Section~\ref{sec:ablation}. The overall reward for the \(g\)-th sample is
\begin{equation}
R^{(g)} = \mathcal{R}_{\text{main}}^{(g)} + \mathcal{R}_{\text{aux}}^{(g)}.
\end{equation}

We then optimize $p_\theta$ with the standard GRPO objective $\mathcal{L}_{\mathrm{GRPO}}$, using group-relative advantages computed from $\{R^{(g)}\}^G_{g=1}$ and a KL penalty against the Stage-1 reference policy.


Following CHORD~\cite{zhang2026onpolicyrlmeetsoffpolicy}, we additionally regularize training with expert trajectories from the first stage. 
The final training loss at optimization step \(s\) is
\begin{equation}
\mathcal{L}^{(s)}
=
(1-\alpha_s)~\mathcal{L}_{\mathrm{GRPO}}
+
\alpha_s\mathcal{L}_{\mathrm{SFT}},
\end{equation}
where \(\alpha_s\) is a step-dependent coefficient as described in CHORD~\cite{zhang2026onpolicyrlmeetsoffpolicy}. Specifically, \(\alpha_s\) gradually decays over the course of training, allowing the regularizing effect of the SFT loss to diminish as the model improves.

%% file: Sections/Experimental-Setup.tex
\section{Experiments}

\input{Figures/TiCo-Bench}
\subsection{TiCo-Bench}
\label{sec:tico-bench}

\noindent \textbf{Benchmark Construction.} Existing spoken dialogue benchmarks evaluate aspects such as response quality, paralinguistic awareness, and turn-taking behavior, but none are designed to measure whether SDMs can follow explicit duration constraints. To fill this gap, we introduce TiCo-Bench, a benchmark dedicated to evaluating time-controllable instruction following in SDMs. 

TiCo-Bench is organized into four task categories, each drawing queries from one
publicly available source: \emph{Question Answering} (\textbf{QA}) from
InstructS2S~\cite{fangllama}, \emph{Reasoning} (\textbf{REA}) from
URO-Bench~\cite{yan2025urobench}, \emph{Creative Generation}
(\textbf{CRE}) from databricks-dolly-15k~\cite{DatabricksBlog2023DollyV2},
and \emph{Summarization} (\textbf{SUM}) from Extreme Summarization
(XSum)~\cite{xsum-emnlp}. 
Figure~\ref{fig:tico-bench-counts} reports the per-category sample counts.
InstructS2S and URO-Bench provide native speech
queries; Dolly-15k and XSum are text only and synthesized via a TTS
pipeline with ASR-based verification
(Appendix~\ref{app:tico-bench-construction}). 
Additionally, Figure~\ref{fig:tico-bench-durations} shows that most inputs lie
below 20\,s, except for SUM which extends to substantially longer durations.


Each base query is paired with an explicit time-control instruction
specifying a target duration, and instantiated under both a \textbf{Short}
setting ($10$--$30$\,s) and a \textbf{Long} setting ($30$--$60$\,s),
giving $2{,}000$ evaluation samples in total. Targets within each regime
are sampled uniformly.

\noindent \textbf{Metrics.} We evaluate duration controllability with Mean 
Absolute Error (MAE, in seconds) and Mean Absolute Percentage Error 
(MAPE, in \%), both computed between the realized speech response duration $d$ and 
the instructed target $t_{\mathrm{inst}}$. MAE captures the absolute 
magnitude of the duration error, while MAPE normalizes by the target 
duration and is thus comparable across target durations.

We further report two quality metrics on a 5-point scale where higher 
is better. \textbf{GPT-score} measures response quality by 
transcribing each generated speech with ASR and prompting 
GPT-5-mini~\cite{openai2025gpt5} to rate the response. \textbf{UTMOS}~\cite{saeki2022utmos} 
measures speech naturalness using a Mean Opinion Score (MOS)
predictor that approximates human MOS ratings.






\noindent\textbf{Baselines.}
We compare against three categories of baselines in TiCo-Bench: (1) open-source SDMs, (2) commercial models, and (3) cascaded systems. For the cascaded strong baselines, we employ an LLM prompted to generate a response that satisfies the target duration constraint as closely as possible, and then use a text-to-speech system to synthesize the corresponding speech. Specifically, we utilize GPT-5.2~\cite{openai2025gpt52systemcard} as a frontier commercial LLM and Qwen2.5-7B-Instruct~\cite{qwen2.5} as a representative SoTA open-source language model. For the TTS component, IndexTTS-2~\cite{zhou2026indextts2} is employed to generate high-quality speech from the LLM response. Detailed prompts  used for the cascaded system can be found at the Appendix~\ref{app:cascade-prompts}.

To ensure that evaluation reflects generation quality rather than truncation artifacts, all SDMs are allocated a sufficiently large token budget to cover responses of up to 1 minute of speech.

\subsection{Experimental Setup}
\label{sec:experimtenal-setup}
We adopted MS-SWIFT~\cite{zhao2024swift}~\footnote{\url{https://github.com/modelscope/ms-swift}} to train the model throughout this paper.
We adopt Qwen-2.5-Omni 7B~\cite{xu2025qwen25} as the backbone model\footnote{The choice of the 7B variant is due to computational constraints.}. Spoken Time Markers are inserted into the output of the ``Thinker''. In both training stages of TiCo, only the ``Thinker'' is trained, while the ``Talker'' remains fixed. During inference, Spoken Time Markers are used only for intermediate planning and are removed via simple regex before feeding the cleaned sequence $\mathbf{z}$ into the ``Talker'' for speech generation.



We sample 4{,}000 speech questions from InstructS2S~\cite{fangllama} as training data, holding out 400 for validation. 
The training data do not overlap with the test set in TiCo-Bench. 
Word-level timestamps for constructing Spoken Time Markers are obtained using Whisper medium~\cite{lintoai2023whispertimestamped}, and a marker is inserted after each sentence-level punctuation mark (e.g., commas, periods, exclamation marks). 
On average each response contains 13.3 markers with a mean inter-marker interval of 2.7 seconds. The full marker distribution is shown in Appendix~\ref{app:training_data_statistics}.


During training, the maximum number of generated tokens for Qwen-2.5-Omni 7B is set to 2{,}048, corresponding to approximately 41 seconds of speech. At inference time, this limit is increased to 4{,}096 to support longer responses. This configuration is primarily adopted for efficiency and to evaluate the model’s ability to generalize to longer outputs, as TiCo-Bench extends up to one minute. In principle, the model can also be trained on longer-response data if desired. Additional training details are provided in the Appendix~\ref{app:training_details}.

%% file: Figures/TiCo-Bench.tex
\begin{figure}[t]
  \centering
  \begin{subfigure}[t]{0.42\linewidth}
    \centering
    \includegraphics[width=\linewidth]{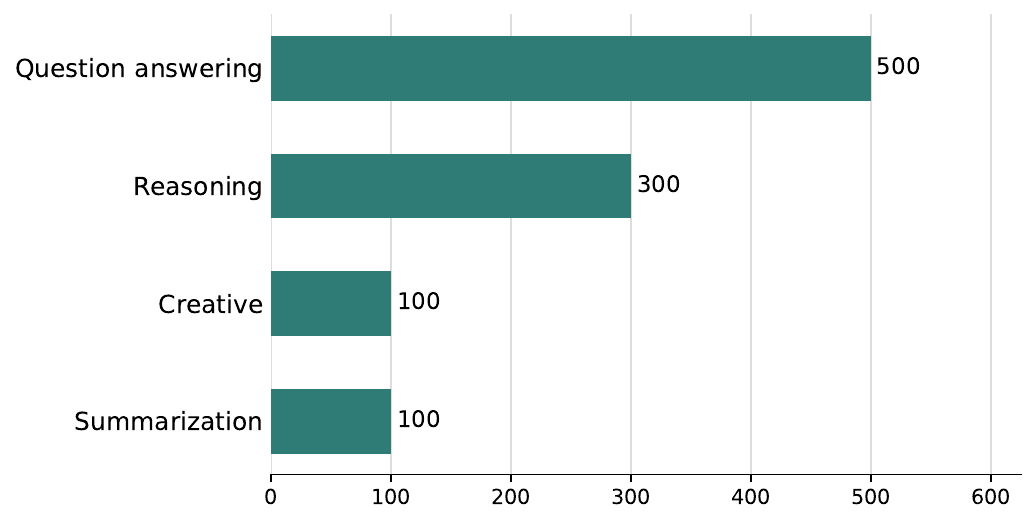}
    \caption{Per-task sample counts.}
    \label{fig:tico-bench-counts}
  \end{subfigure}\hfill
  \begin{subfigure}[t]{0.56\linewidth}
    \centering
    \includegraphics[width=\linewidth]{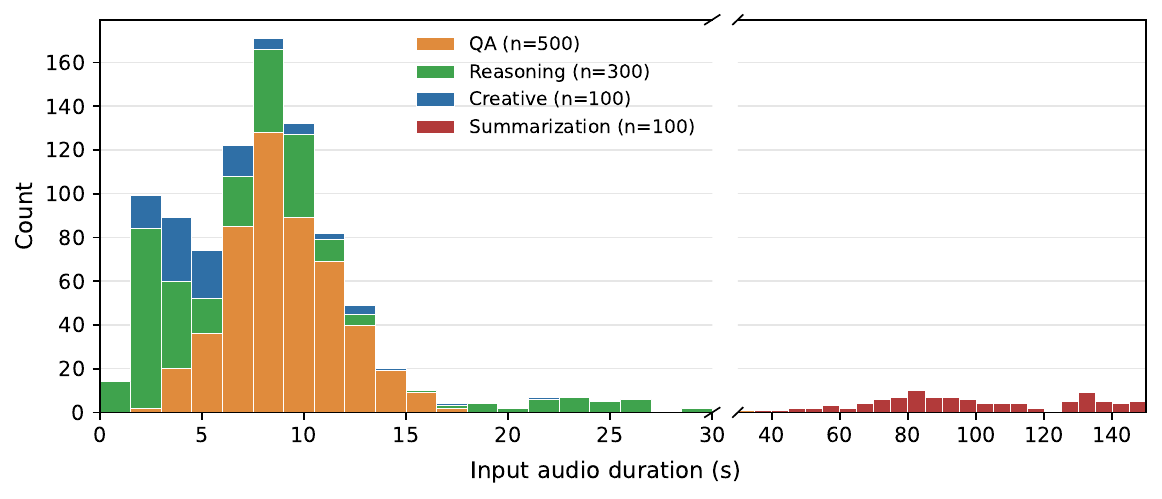}
    \caption{Input-audio duration distribution per task.}
    \label{fig:tico-bench-durations}
  \end{subfigure}
  \caption{
      Composition of \textbf{TiCo-Bench}.
      (a) The benchmark contains 1{,}000 base speech queries spanning four task
      categories: Question Answering, Reasoning, Creative, and
      Summarization.
      (b) Most queries fall within a short duration range, while only Summarization extends to substantially longer durations.
  }
  \label{fig:tico-bench-composition}
\end{figure}

%% file: Sections/Results.tex
\section{Results}
\subsection{TiCo-Bench}
\label{sec:tico-bench-performance}

Table~\ref{tab:speech_q_results_mape} reports our main results on 
TiCo-Bench. TiCo achieves the lowest MAPE on 7 out of 8 tasks with an 
overall error of 16.2\%, a $2.7\times$ reduction over its backbone 
Qwen2.5-Omni-7B (43.3\%) and a $1.6\times$ reduction over the strongest 
baseline Cascade~(GPT) (25.2\%). The gain is uniform across both 
duration regimes and various tasks, suggesting a generic time-aware planning capability rather than 
task-specific adaptation. 
The single exception is Short–SUM, where both Cascade (GPT) (17.1\%) and Qwen3-Omni-30B (32.2\%) outperform TiCo (49.0\%). This long-input, short-output regime is 
rare in our Stage 1 self-generation data, where responses tend to be longer than the input queries.
Notably, Qwen3-Omni-30B, despite being roughly $4\times$ larger than Qwen2.5-Omni-7B, achieves an overall MAPE of 42.1\%, only marginally better than Qwen2.5-Omni-7B (43.3\%). This indicates that duration controllability does not naturally emerge from larger models, 
motivating our effective post-training method for time controllability.

End-to-end SDMs and GPT-audio incur substantially larger \emph{relative} 
error in the Short setting than the Long setting, mirroring the verbosity 
bias documented for text LLMs~\cite{dubois2024length, hu-etal-2025-explaining}. Cascaded systems suppress 
this bias by planning duration in text, but the realized speech duration 
depends on the downstream TTS speaking rate, which is unobservable to the 
planning LLM. The two failure modes are dual: SDMs control content but not 
realized timing, while cascaded systems control intended timing but not 
realized speech. The Short--QA vs.\ Long--QA contrast illustrates this: 
Cascade~(GPT) MAPE \emph{rises} from 19.9\% to 24.3\%, whereas TiCo 
\emph{decreases} from 15.4\% to 11.7\%. Spoken Time Markers expose the 
model's own realized speaking time inside the generation loop, which we view 
as the mechanistic reason TiCo avoids both trade-offs.

TiCo preserves response quality (GPT-score: 3.32 vs.\ backbone 3.31) and 
speech naturalness (UTMOS 4.04 vs.\ 4.09; ground-truth 4.08), consistent 
with Spoken Time Markers being stripped before speech synthesis. This rules 
out two alternative explanations for the MAPE gain: reward hacking at the 
cost of content, and distortion from the inserted markers.

\input{Tables/main_results_v2}



\subsection{Generalization to Longer Responses and Text Queries}
We further examine whether TiCo generalizes beyond the conditions seen during post-training.

\input{Figures/fig_speech}

As shown in Figure~\ref{fig:speech-vs-target}, despite being post-trained with responses of at most 41 seconds, TiCo maintains consistently low MAE and MAPE across instructed-duration bins on both InstructS2S and UROBench, with relative error on long-duration bins comparable to or even lower than that on short-duration bins, while the backbone's error grows noticeably for targets beyond 45 seconds.

We further examine generalization across input modalities by evaluating on two additional benchmarks with text queries. TiCo achieves an overall MAPE of 18.0\%, a 1.6$\times$ reduction over the strongest baseline Cascade (GPT) and a 2.7$\times$ reduction over the backbone, comparable to its 16.2\% MAPE on speech queries despite being trained solely on speech. Response quality is preserved under the modality shift, and the Spoken Time Markers continue to faithfully track realized speaking time on textual inputs. Detailed results are reported in Appendix~\ref{app:text-generalization}.



\subsection{Spoken Time Marker Prediction Analysis}
\label{sec:stm-analysis}

\input{Figures/fig_speech_timemarker}

Table~\ref{tab:speech_q_results_mape} establishes that TiCo controls 
duration well. To verify that this comes from the markers genuinely 
tracking the realized speaking time, and thus serving as a real-time 
planning signal, we test marker accuracy at two granularities. The first is global 
accuracy, asking whether the final marker at the end of the response 
matches the realized response duration. 
Figure~\ref{fig:speech-vs-token} compares the duration error against 
the instructed duration with the error against the final marker. The 
two curves track each other closely across all instructed-duration 
bins on both InstructS2S and URO-Bench, with a small and roughly 
constant gap indicating an additive offset rather than scale-dependent 
error. The second is local accuracy, asking whether each intermediate 
marker matches the realized speaking time at its position. Aligning 
markers with Whisper word-level timestamps yields local marker errors 
averaging 2.65 seconds under Short and 3.21 seconds under Long 
settings (per-task breakdown in Appendix~\ref{app:stm-perword}), 
confirming that markers can serve as a real-time planning signal 
during generation.

\subsection{RL Reward Ablation}
\label{sec:ablation}
Table~\ref{tab:ablation} ablates the reward components used in the second-stage Time-Controllable Training with GRPO. Models are evaluated on a 720-sample subset of TiCo-Bench. With only the main duration reward ($\mathcal{R}^{(g)}_{\text{main}}$), GRPO already outperforms the backbone model but exhibits reward hacking.
Adding $\mathcal{R}^{(g)}_{\text{pres}}$ alone hurts performance, but combining it with $\mathcal{R}^{(g)}_{\text{mono}}$ recovers the baseline and improves MAE to 7.66s, showing that monotonicity is critical. Substituting presence with $\mathcal{R}^{(g)}_{\text{copy}}$ further reduces MAE to 5.30s, and adding $\mathcal{R}^{(g)}_{\text{rep}}$ brings it down to 4.71s. The \emph{Full} configuration, integrating all components, achieves the best performance (MAE 4.55s, MAPE 15.38\%), demonstrating that the reward components are complementary and jointly necessary.
Overall, the main reward alone is sufficient to improve time-controllability over the base model, while the auxiliary rewards further prevent reward hacking and yield the best performance when combined.

\input{Tables/reward_ablation}

%% file: Tables/main_results_v2.tex
\begin{table*}[t]
\centering
\caption{
\textbf{TiCo-Bench} evaluation on \textbf{speech-query} tasks 
under Short (10--30\,s) and Long (30--60\,s) settings, broken down by task 
category (QA, REA, CRE, SUM). Per-task and overall scores are MAPE (\%); 
rightmost columns report GPT-score and UTMOS averaged 
over all subsets. Model categories: 
\legenddot{cascadedcolor}~Cascaded, 
\legenddot{commercialcolor}~Commercial, 
\legenddot{opensourcecolor}~Open-sourced, 
\legenddot{proposedcolor}~Proposed. 
\textbf{Bold} marks the best result per column.
$^{*}$Kimi-Audio SUM clips exceeding the model's 120\,s context are excluded. 
Per-task MAE, GPT-score, and UTMOS performance are in 
Appendix~\ref{app:detailed-results}.
}
\label{tab:speech_q_results_mape}
\renewcommand{\arraystretch}{1.15}
\resizebox{\textwidth}{!}{%
\begin{tabular}{>{\raggedright\arraybackslash}m{3.4cm} cccc cccc ccc}
\toprule
\emph{TiCo-Bench (Speech)}
& \multicolumn{4}{c}{Short (10s--30s)}
& \multicolumn{4}{c}{Long (30s--60s)}
& \multicolumn{3}{c}{Overall} \\
\cmidrule(lr){2-5} \cmidrule(lr){6-9} \cmidrule(lr){10-12}
\textbf{Model}
& \textbf{QA} & \textbf{REA} & \textbf{CRE} & \textbf{SUM}
& \textbf{QA} & \textbf{REA} & \textbf{CRE} & \textbf{SUM}
& \textbf{MAPE $\downarrow$} & \textbf{GPT-score $\uparrow$} & \textbf{UTMOS $\uparrow$} \\
\midrule

\cascadedbox{Cascade (Qwen)}
& 36.52 & 51.43 & 35.93 & 59.19
& 46.32 & 65.32 & 38.90 & 24.89
& 46.17 & 3.18 & 3.00 \\

\cascadedbox{Cascade (GPT)}
& 19.90 & 28.07 & 19.97 & \textbf{17.10}
& 24.28 & 39.55 & 25.81 & 16.84
& 25.17 & \textbf{3.92} & 2.91 \\

\commercialbox{GPT-audio}
& 82.53 & 107.91 & 127.97 & 58.37
& 45.96 & 48.21 & 49.51 & 41.01
& 69.38 & 3.45 & 3.72 \\

\opensourcebox{Kimi Audio}
& 133.17 & 166.79 & 156.45 & 60.65$^{*}$
& 60.42 & 72.43 & 59.66 & 46.18$^{*}$
& 101.82 & 1.69 & 2.43 \\

\opensourcebox{LFM Audio}
& 89.13 & 61.46 & 73.46 & 48.13
& 47.66 & 74.02 & 53.99 & 48.07
& 65.72 & 2.58 & 4.01 \\

\opensourcebox{MiMo-Audio}
& 74.74 & 75.06 & 123.27 & 68.20
& 45.80 & 38.18 & 40.92 & 31.31
& 60.31 & 2.91 & 3.31 \\

\opensourcebox{Qwen3-Omni-30B}
& 51.65 & 50.16 & 75.86 & 32.19
& 30.00 & 38.26 & 25.05 & 34.53
& 42.06 & 3.58 & \textbf{4.44} \\
\midrule

\opensourcebox{Qwen2.5-Omni-7B}
& 49.61 & 45.97 & 55.99 & 79.55
& 32.94 & 42.13 & 29.82 & 24.93
& 43.33 & 3.31 & 4.09 \\

\proposedbox{\textbf{TiCo (Proposed)}}
& \textbf{15.44} & \textbf{17.81} & \textbf{17.32} & 48.96
& \textbf{11.66} & \textbf{13.92} & \textbf{13.07} & \textbf{13.51}
& \textbf{16.19} & 3.32 & 4.04 \\

\bottomrule
\end{tabular}
}
\end{table*}

%% file: Figures/fig_speech.tex

\begin{figure*}[h]
    \centering
    \includegraphics[width=\textwidth]{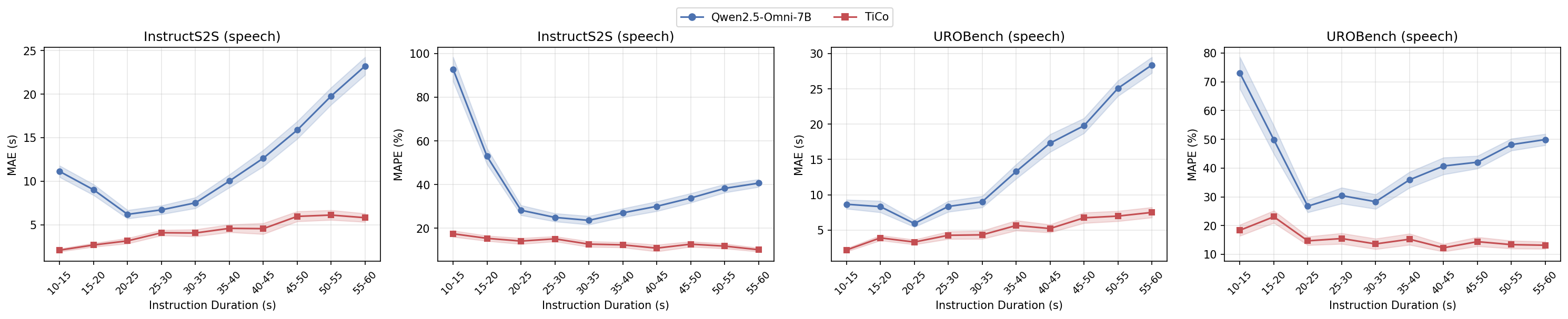}
    \caption{Duration MAE and MAPE of Qwen2.5-Omni-7B and TiCo across instructed-duration bins on two speech benchmarks, InstructS2S and UROBench. TiCo maintains consistently low error across the full duration range, while the backbone's error grows for targets beyond 45 seconds.}
    \label{fig:speech-vs-target}
\end{figure*}


%% file: Figures/fig_speech_timemarker.tex
\begin{figure*}[h!]
    \centering
    \includegraphics[width=\textwidth]{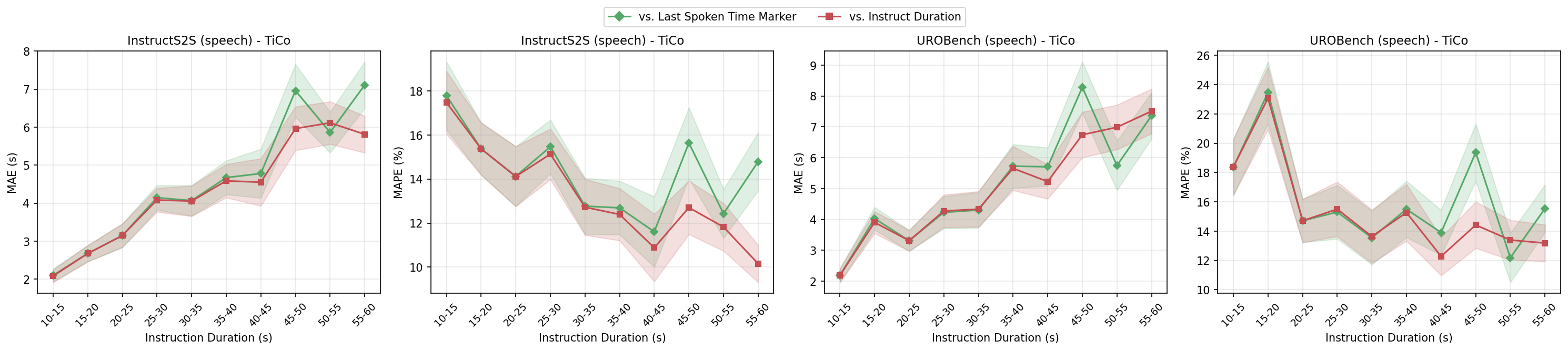}
    \caption{Duration error of TiCo across instructed-duration bins, comparing two reference signals: the instructed duration $t_{\mathrm{inst}}$ and the final Spoken Time Marker $t_{\mathrm{last}}$. The close alignment indicates that the final time marker accurately estimates realized speech duration.}
    \label{fig:speech-vs-token}
\end{figure*}

%% file: Tables/reward_ablation.tex
\begin{table}[h]
\centering
\caption{Ablation of reward components in second-stage training on a TiCo-Bench subset.}
\resizebox{0.9\linewidth}{!}{%
\begin{tabular}{lc|cccccc}
\toprule
\textbf{Variant} & $\mathcal{R}^{(g)}_{\text{main}}$ & $\mathcal{R}^{(g)}_{\text{pres}}$ & $\mathcal{R}^{(g)}_{\text{mono}}$ & $\mathcal{R}^{(g)}_{\text{copy}}$ & $\mathcal{R}^{(g)}_{\text{rep}}$ & \textbf{MAE (s)} & \textbf{MAPE (\%)} \\
\midrule
\textit{Main only} & \checkmark & & & & & 8.39 & 32.21 \\
\textit{+ Presence} & \checkmark & \checkmark & & & & 10.18 & 38.89 \\
\textit{+ Presence + Mono.} & \checkmark & \checkmark & \checkmark & & & 7.66 & 28.10 \\
\textit{+ Mono. + Copy} & \checkmark & & \checkmark & \checkmark & & 5.30 & 19.07 \\
\textit{+ Mono. + Copy + Rep.} & \checkmark & & \checkmark & \checkmark & \checkmark & 4.71 & 17.16 \\
\midrule
\textit{Full} & \checkmark & \checkmark & \checkmark & \checkmark & \checkmark & \textbf{4.55} & \textbf{15.38} \\
\bottomrule
\end{tabular}
}
\label{tab:ablation}
\end{table}

%% file: Sections/Conclusion.tex
\section{Conclusion}
We introduced \textbf{TiCo}, a time-controllable spoken dialogue model that follows explicit duration constraints by exposing realized speaking time inside the generation loop through Spoken Time Markers. TiCo is obtained through a two-stage post-training procedure. The first stage leverages self-generation to instill time awareness, and the second stage applies reinforcement learning with verifiable duration rewards to sharpen controllability while preserving response quality. To support systematic evaluation, we further introduced \textbf{TiCo-Bench}, the first benchmark for time-controllable instruction following in spoken dialogue models. On TiCo-Bench, TiCo reduces duration error by $2.7\times$ over its backbone and $1.6\times$ over the strongest baseline, with response quality and speech naturalness remaining comparable to the backbone. The capability further generalizes beyond the duration range seen during training and transfers from speech to text queries, suggesting that temporal control can be acquired as a robust intermediate planning skill rather than a task-specific behavior. 


%% file: Sections/Appendix.tex
\newcommand{\ts}[1]{\mbox{\texttt{\textless #1 seconds\textgreater}}}

\newtcolorbox{responseboxblue}[1]{
  breakable,
  enhanced,
  colback=blue!4,
  colframe=blue!45,
  boxrule=0.4pt,
  arc=1mm,
  left=1.2mm,
  right=1.2mm,
  top=0.8mm,
  bottom=0.8mm,
  title=\textbf{#1}
}

\newtcolorbox{responseboxgreen}[1]{
  breakable,
  enhanced,
  colback=green!4,
  colframe=green!45!black,
  boxrule=0.4pt,
  arc=1mm,
  left=1.2mm,
  right=1.2mm,
  top=0.8mm,
  bottom=0.8mm,
  title=\textbf{#1}
}

\newtcolorbox{responseboxgray}[1]{
  breakable,
  enhanced,
  colback=gray!5!white,
  colframe=gray,
  boxrule=0.4pt,
  arc=1mm,
  left=1.2mm,
  right=1.2mm,
  top=0.8mm,
  bottom=0.8mm,
  title=\textbf{#1}
}






\section{Acknowledgment}
This work was supported by the Ministry of Education (MOE) of Taiwan under the project Taiwan Centers of Excellence in Artificial Intelligence, through the NTU Artificial Intelligence Center of Research Excellence (NTU AI-CoRE).

\input{Sections/Limitation}

\section{Training Details}
\label{app:training_details}
\noindent \textbf{Stage 1: Time-Awareness SFT.}
We fine-tune Qwen2.5-Omni-7B with LoRA ($r{=}8$, $\alpha{=}16$) on all linear layers, keeping the vision encoder frozen.
The training set consists of 4{,}000 samples (400 held out for validation).
We train for 5 epochs with a batch size of 2 per GPU $\times$ 4 GPUs and gradient accumulation of 4 steps (effective batch size 32), using a cosine learning rate schedule with peak $5 \times 10^{-5}$ and 10\% warmup.
Maximum sequence length is 1{,}024 tokens.
Training uses bfloat16 precision with gradient checkpointing.

\noindent \textbf{Stage 2: Time-Controllable GRPO with CHORD.}
Starting from the Stage~1 checkpoint, we apply GRPO with CHORD~\cite{zhang2026onpolicyrlmeetsoffpolicy} to optimize duration controllability.
The LoRA configuration uses $r{=}8$, $\alpha{=}32$.
We train for 800 steps with a per-GPU batch size of 1 and gradient accumulation of 8 (effective batch size 32), learning rate $5 \times 10^{-6}$ with cosine schedule and 10\% warmup.
Each prompt generates $G{=}4$ candidate completions with maximum completion length of 512 tokens. The clipping parameter is $\varepsilon{=}0.2$ and the KL penalty coefficient is $\beta{=}0.04$.

\noindent \textbf{Reward Design.}
The main reward function is 
\begin{equation}
\mathcal{R}_{\text{main}}^{(g)}
= F \left(t_{\text{inst}} - t_{\text{last}}^{(g)}\right),
\end{equation}
where \(t_{\text{inst}}\) is the target duration specified in the instruction and \(t_{\text{last}}^{(g)}\) is the duration indicated by the final generated time marker in \(\tilde{\mathbf{z}}^{(g)}\).
The function \(F\) is defined as a Gaussian:
\begin{equation}
F(\Delta t) = \exp\!\left(-\frac{(\Delta t)^2}{2\sigma^2}\right),
\end{equation}
where \(\sigma\) controls the tolerance to duration deviations. In our experiments, we set \(\sigma = 5\).

We further incorporate auxiliary reward functions to prevent reward hacking:
\begin{itemize}
    \item \textbf{Presence reward} $\mathcal{R}_{\text{pres}}^{(g)}$: encourages the model to generate at least one Spoken Time Marker,
    \begin{equation}
    \mathcal{R}_{\text{pres}}^{(g)} = \mathbb{I}\!\left[\,M_g \geq 1\,\right],
    \end{equation}
    where $M_g$ denotes the number of time markers in $\tilde{\mathbf{z}}^{(g)}$.

    \item \textbf{Monotonicity reward} $\mathcal{R}_{\text{mono}}^{(g)}$: encourages generated time markers to be strictly increasing. We compute the fraction of consecutive pairs that are strictly increasing:
    \begin{equation}
    \mathcal{R}_{\text{mono}}^{(g)}
    =
    \frac{1}{M_g - 1}
    \sum_{j=1}^{M_g - 1}
    \mathbb{I}\!\left[\,t_{j+1}^{(g)} > t_j^{(g)}\,\right].
    \end{equation}

    \item \textbf{Repetition penalty} $\mathcal{R}_{\text{rep}}^{(g)}$: penalizes repeated time marker values:
    \begin{equation}
    \mathcal{R}_{\text{rep}}^{(g)}
    =
    -\left(1 - \frac{|\{t_1^{(g)}, \dots, t_{M_g}^{(g)}\}|}{M_g}\right),
    \end{equation}
    where $|\cdot|$ denotes set cardinality. The penalty is $0$ when all markers are unique and $-1$ when all are identical. 

    \item \textbf{Copy penalty} $\mathcal{R}_{\text{copy}}^{(g)}$: penalizes non-final time markers that trivially copy the instructed duration $t_{\text{inst}}$:
    \begin{equation}
    \mathcal{R}_{\text{copy}}^{(g)}
    =
    -\frac{1}{M_g}
    \sum_{j=1}^{M_g - 1}
    \mathbb{I}\!\left[\,|t_j^{(g)} - t_{\text{inst}}| < \tau\,\right],
    \end{equation}
    where $\tau{=}0.5$\,s is the tolerance threshold. The final marker $t_{M_g}^{(g)}$ is excluded since matching the target duration at the end is the desired behavior. 
\end{itemize}

The overall reward for the $g$-th sample is
\begin{equation}
R^{(g)}
=
\mathcal{R}_{\text{main}}^{(g)}
+
\mathcal{R}_{\text{pres}}^{(g)}
+
\mathcal{R}_{\text{mono}}^{(g)}
+
\mathcal{R}_{\text{rep}}^{(g)}
+
\mathcal{R}_{\text{copy}}^{(g)}.
\end{equation}
Note that $\mathcal{R}_{\text{rep}}^{(g)}$ and $\mathcal{R}_{\text{copy}}^{(g)}$ are non-positive by construction, so no explicit subtraction is needed.

\noindent \textbf{CHORD.}
CHORD interleaves SFT updates with GRPO updates using a mixing coefficient $\mu$ that decays from $\mu_{\text{peak}}{=}0.8$ to $\mu_{\text{valley}}{=}0.3$ over 500 steps, preventing catastrophic forgetting of general conversational ability.
Both stages are trained on 4 NVIDIA A6000 GPUs, and the entire two-stage pipeline completes in less than one day.

\noindent \textbf{Training Data Statistics.} Figure~\ref{fig:time-marker-stats} shows the distribution of Spoken 
Time Markers in the Stage 1 self-generated training data. Each 
response contains on average 13.3 markers, with a mean inter-marker 
interval of 2.70 seconds. The marker timestamps are most densely 
distributed within the first 20--30 seconds, reflecting the typical 
length range of self-generated responses.

\label{app:training_data_statistics}
\input{Figures/time_marker/time_marker}

\section{Cascaded System Prompt Templates}
\label{app:cascade-prompts}

We use a unified system prompt across GPT and Qwen for the cascaded LLM baseline.

\begin{responseboxgray}{LLM System Prompt for Cascaded System}

\emph{You are writing a final script for text-to-speech (TTS). Your response will be synthesized directly into speech. Follow the duration instruction as strictly as possible. Output only the final spoken text, with natural punctuation. Do not output markdown, bullets, JSON, XML tags, stage directions, or extra commentary. Do not mention these instructions.}

\end{responseboxgray}






\section{TiCo-Bench Construction Details}
\label{app:tico-bench-construction}

This appendix expands on the construction of TiCo-Bench summarized in
Section~\ref{sec:tico-bench}. We describe the source datasets and their
licenses, the TTS pipeline used to obtain spoken queries for the text-only
sources, and the time-control instruction sampling protocol.

\subsection{Source Datasets}
\label{app:bench-sources}

TiCo-Bench draws speech queries from four publicly available sources, one per
task category. We describe each below, including the rationale for assigning
it to its category and the license under which the source dataset is
distributed.

\paragraph{Question Answering (QA): InstructS2S~\cite{fangllama}.}
InstructS2S is the spoken instruction-following corpus released alongside
LLaMA-Omni, in which information-seeking questions are paired with
single-voice speech recordings. Its prompts are predominantly factual
question-answering, which directly matches the QA category. We sample 500
unique English queries that do not overlap with our training subset.
The dataset is released under the CC~BY-NC 4.0 license.

\paragraph{Reasoning (REA): URO-Bench~\cite{yan2025urobench}.}
URO-Bench is a comprehensive evaluation suite for end-to-end spoken dialogue
models, organized into capability-specific subsets. We restrict our sampling
to its reasoning-oriented subsets, which together cover narrative reasoning,
truthfulness, mathematical reasoning, multi-domain knowledge, and open-ended
multi-turn reasoning. Drawing from these subsets ensures that the REA
category in TiCo-Bench targets reasoning ability rather than generic
question answering. We sample 300 queries in total. URO-Bench is released
under the MIT license.

\paragraph{Creative Generation (CRE): databricks-dolly-15k Creative Writing~\cite{DatabricksBlog2023DollyV2}.}
databricks-dolly-15k is an open-source instruction-tuning dataset organized
into eight task categories. We use only its \texttt{creative\_writing}
subset, in which prompts ask the respondent to produce open-ended creative
content such as a short rant, an imaginary monologue, or a stylized rewrite.
The open-ended nature of these prompts makes the natural response length
flexible, which is the property required of the CRE category. The dataset
is released under the CC~BY-SA 3.0 license.

\paragraph{Summarization (SUM): Extreme Summarization (XSum)~\cite{xsum-emnlp}.}
XSum pairs each BBC news article with a single professionally written summary
sentence. We use the article side as the input and prompt the model to
produce a spoken summary under a target duration. Articles are substantially
longer than the queries in the other three categories, which provides a natural source for the long-input regime that 
distinguishes the SUM category from the others. The dataset is released under the MIT license.

\textbf{All four source datasets are released under licenses that permit 
redistribution and adaptation for non-commercial research, allowing 
us to include their queries as part of TiCo-Bench.}

\subsection{TTS Pipeline and Quality Control}

For the Dolly-15k and XSum sources, which are text only, we obtain spoken
queries through a TTS-then-verify pipeline. Each original instruction is
synthesized with CosyVoice~3~\cite{du2025cosyvoice}, using reference audio
randomly sampled from LibriTTS~\cite{Zen2019-kz} as the prompt voice and
applying no paralinguistic conditioning. Synthesized waveforms are produced
at 24~kHz mono.

The two text sources are verified with different procedures, reflecting their
distinct failure modes. For XSum, each synthesized utterance is transcribed
using Whisper with the large-v3 checkpoint and compared against the
source article using the \texttt{jiwer} library, after applying the Whisper
English text normalizer to both sides. An utterance is retained only if its
word error rate is below 0.1 and its duration falls within $[30, 150]$
seconds; the duration filter is what gives the SUM category its long-input
regime. For Dolly, ASR-based verification is unreliable because
creative-writing prompts contain rare proper nouns and unconventional
phrasing that the ASR can mistranscribe even when the synthesized speech is
intelligible. We therefore manually review every synthesized utterance and
retain only those that are intelligible and faithful to the original prompt.

\subsection{Time-Control Instruction Sampling}

For each base query we sample a target duration $t_{\mathrm{inst}}$ uniformly
from $[10, 30]$ seconds in the Short setting and from $(30, 60]$ seconds in
the Long setting. Targets are quantized to one-second granularity, and the
two regimes use disjoint target ranges. The two settings therefore give
$1{,}000 \times 2 = 2{,}000$ evaluation samples in total.

The textual time-control instruction is instantiated from a pool of
semantically equivalent templates, with one template sampled uniformly per
query. The instruction is delivered to the model as a textual
turn alongside the speech query, following the conversation format of each
model under evaluation.

\section{Generalization Study of Textual Queries}
\label{app:text-generalization}

During both training stages of TiCo, the model always receives 
speech queries as input. Here we evaluate whether the trained model generalizes to text queries on two benchmarks.

The first benchmark is \textbf{LIFEBench}~\cite{zhang2025lifebench}, 
a length-instruction benchmark for text LLMs in which each query is 
paired with a word-count target. We extract its English queries and 
replace the original word-count instruction with a time-control 
instruction under the same Short (10--30\,s) and Long (30--60\,s) 
regimes as TiCo-Bench.

The second benchmark uses the textual transcriptions provided by 
\textbf{URO-Bench}~\cite{yan2025urobench}. URO-Bench is a speech 
benchmark that pairs each speech query with its textual 
transcription. We use the transcriptions as text input, drawing 
from the same reasoning subsets as the REA category in TiCo-Bench, 
though with a smaller sample size. Targets follow the same Short 
and Long protocol as in TiCo-Bench.

Table~\ref{tab:text_q_results} reports the time-control performance 
on text queries. TiCo achieves the lowest MAE and MAPE on all four 
benchmark-regime cells, with an overall MAPE of 18.0\%, a 
1.6$\times$ reduction over the strongest baseline Cascade (GPT) 
(28.1\%) and a 2.7$\times$ reduction over the backbone 
Qwen2.5-Omni-7B (48.9\%). Notably, this text-query performance is 
comparable to TiCo's speech-query MAPE on TiCo-Bench (16.2\%, 
Table~\ref{tab:speech_q_results_mape}), despite the model being 
trained exclusively on speech queries. As shown in 
Figure~\ref{fig:text-vs-target}, TiCo maintains consistently low 
error across all instructed-duration bins on both LIFEBench and 
UROBench-text, while the backbone's error grows substantially for 
longer targets. The GPT-score (2.76 vs.\ backbone 2.67) further 
confirms that response quality is preserved under the modality 
shift.

Figure~\ref{fig:text-vs-token} verifies that the marker-based 
planning mechanism remains faithful on text queries. The error of 
TiCo's outputs against the instructed duration tracks closely the 
error against the final Spoken Time Marker on both benchmarks, 
with a small and roughly constant gap. This mirrors the 
speech-query finding in Section~\ref{sec:stm-analysis}, indicating 
that the markers continue to track realized speaking time even 
when inputs are textual rather than spoken.

\input{Tables/main_results_text}
\input{Figures/fig_text}
\FloatBarrier

\section{Detailed Per-Subset Results on TiCo-Bench}
\label{app:detailed-results}

\input{Tables/per_task_mae}

\input{Tables/per_task_gptscore}
\input{Tables/per_task_utmos}
\FloatBarrier

\section{Local Alignment of Spoken Time Marker}
\label{app:stm-perword}

This appendix provides the per-task breakdown of the local-accuracy 
analysis summarized in Section~\ref{sec:stm-analysis}. To assess 
marker accuracy at every position rather than only at the end of the 
response, we align each Spoken Time Marker in the generated text with 
the word-level timestamp obtained from a Whisper-based ASR pass over 
the synthesized speech. For each matched position we compute the 
absolute error in seconds between the predicted marker time and the 
ASR-aligned word timestamp, then average across positions and samples 
within each task and duration regime. 
Table~\ref{tab:stm-alignment} reports the resulting local marker errors.

\input{Tables/timemarker_mae}

\section{Qualitative Examples}
\label{app:qualitative_examples}
\input{Tables/qualitative}

\section{Spoken Dialogue Model Survey}
\label{Appendix:Survey}

\input{Figures/pattern.drawio}

Table~\ref{tab:sdm_ir_survey} surveys representative spoken dialogue models (SDMs), including their Intermediate Representations \textbf{(IR)}, target speech representations \textbf{(Speech Rrep.)}, and the generation patterns \textbf{(Pattern)} that describe how intermediate representations and speech representations are processed during speech response generation.  Readers may refer to the spoken language model (SLM) survey paper~\cite{aroralandscape} for a more detailed discussion on speech representation and generation pattern.

\noindent \textbf{Intermediate Representation (IR).}
With the emergence of text-based large language models (LLMs) demonstrating strong reasoning capabilities, modern spoken dialogue models (SDMs) increasingly adopt LLMs to generate speech responses, using text as an intermediate representation for semantic planning. 

Text-based IR offers high versatility and can serve multiple purposes, as summarized in Table~\ref{tab:sdm_ir_survey}, including style control (\(+\)S.), reasoning (\(+\)R.), tool calling (\(+\)Tool), and direct guidance of the target speech content.

\noindent \textbf{Pattern.}
The intermediate representation and the target speech representations can be generated under several design patterns, each leading to different trade-offs in terms of efficiency, latency, and the degree to which speech generation is conditioned on the intermediate representation.
For simplicity, we assume text as the intermediate representation in the following discussion, and provide an illustration in Figure~\ref{fig:Pattern-SDM}.

\begin{itemize}
    
\item \textbf{Sequential}: Text is generated first, followed by speech tokens. Chunking strategies can be incorporated to support streaming generation.

\item \textbf{Parallel}: Text and speech tokens are generated simultaneously. In this setting, the hidden representations of a text LLM are typically used to predict text tokens and speech tokens through separate prediction networks. Frame-level operations can further be introduced to realize delay patterns.

\item \textbf{Interleaved}: Text and speech tokens are arranged in a single interleaved sequence, typically modeled by a single LLM, allowing speech representations to be conditioned more directly on text representations.
\end{itemize}

\noindent \textbf{Speech Representations.}
The goal of a spoken dialogue model is to generate an appropriate spoken response, typically represented as a sequence of speech tokens. These tokens can be further synthesized into waveforms using a pre-trained vocoder or an audio codec decoder.

\begin{itemize}
\item \textbf{Phonetic tokens.}
Phonetic tokens are obtained by quantizing speech encoder representations (e.g., via K-means), such as those extracted from self-supervised speech models (e.g., HuBERT) or foundation ASR models (e.g., Whisper encoders). 
They primarily capture phonetic and linguistic content, while containing relatively limited acoustic information such as speaker identity or environmental characteristics. 
In prior work, they are also referred to as \emph{semantic tokens}~\cite{borsos2023audiolm}.

When \emph{phonetic tokens} are used as the target speech representation, an additional vocoder (e.g., HiFi-GAN or flow-matching decoders) is typically required to incorporate speaker identity and speaking style, as these attributes are not explicitly encoded.

\item \textbf{Acoustic tokens.}
Acoustic tokens~\cite{mousavi2025discrete, guo2025recent, wu2024codec} are derived from neural speech codec models trained with reconstruction objectives. 
These models typically employ multiple hierarchical codebooks based on residual vector quantization (RVQ).

When \emph{acoustic tokens} are generated, a pre-trained audio codec decoder can be directly used for waveform synthesis.
Recently, there has been a growing trend toward distilling phonetic information into the early layers of acoustic tokens, aiming to preserve phonetic structure while maintaining rich acoustic detail~\cite{zhangspeechtokenizer, defossez2024moshi}.
\end{itemize}

\emph{(Please find the survey table on the following page.)}

\input{Tables/survey}

%% file: Sections/Limitation.tex
\section{Limitations}
\label{app:limitations}
While TiCo significantly improves duration controllability, several 
limitations remain. First, TiCo underperforms Cascade (GPT) on the 
Short--SUM subset, where the long-input, short-output regime is 
under-represented in our Stage 1 self-generation training data. 
Second, the local marker prediction error remains around 2--3 
seconds, which limits the precision of fine-grained intermediate 
planning. Third, our experiments are conducted on a single backbone 
(Qwen2.5-Omni 7B) with the Thinker-Talker architecture, and whether 
the Spoken Time Marker mechanism transfers to spoken dialogue models 
with parallel or interleaved generation patterns remains untested. 
Finally, our training data is drawn solely from English 
instruction-following queries, and generalization to other domains and conversational settings has not yet been evaluated.

%% file: Figures/time_marker/time_marker.tex
\begin{figure*}[h]
    \centering
    \begin{subfigure}[t]{0.32\textwidth}
        \centering
        \includegraphics[width=\textwidth]{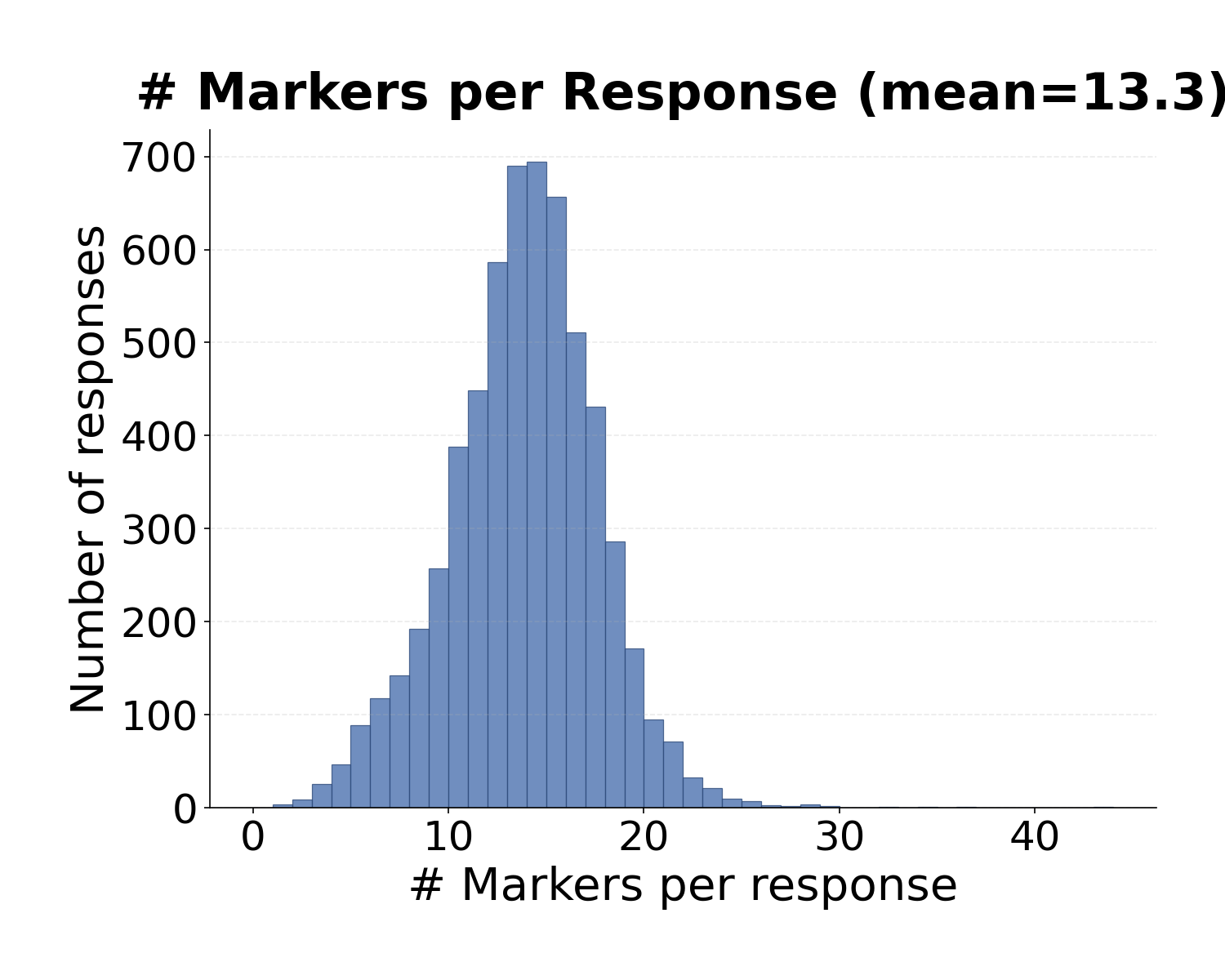}
        \caption{Markers per Response}
        \label{fig:tm-counts}
    \end{subfigure}
    \hfill
    \begin{subfigure}[t]{0.32\textwidth}
        \centering
        \includegraphics[width=\textwidth]{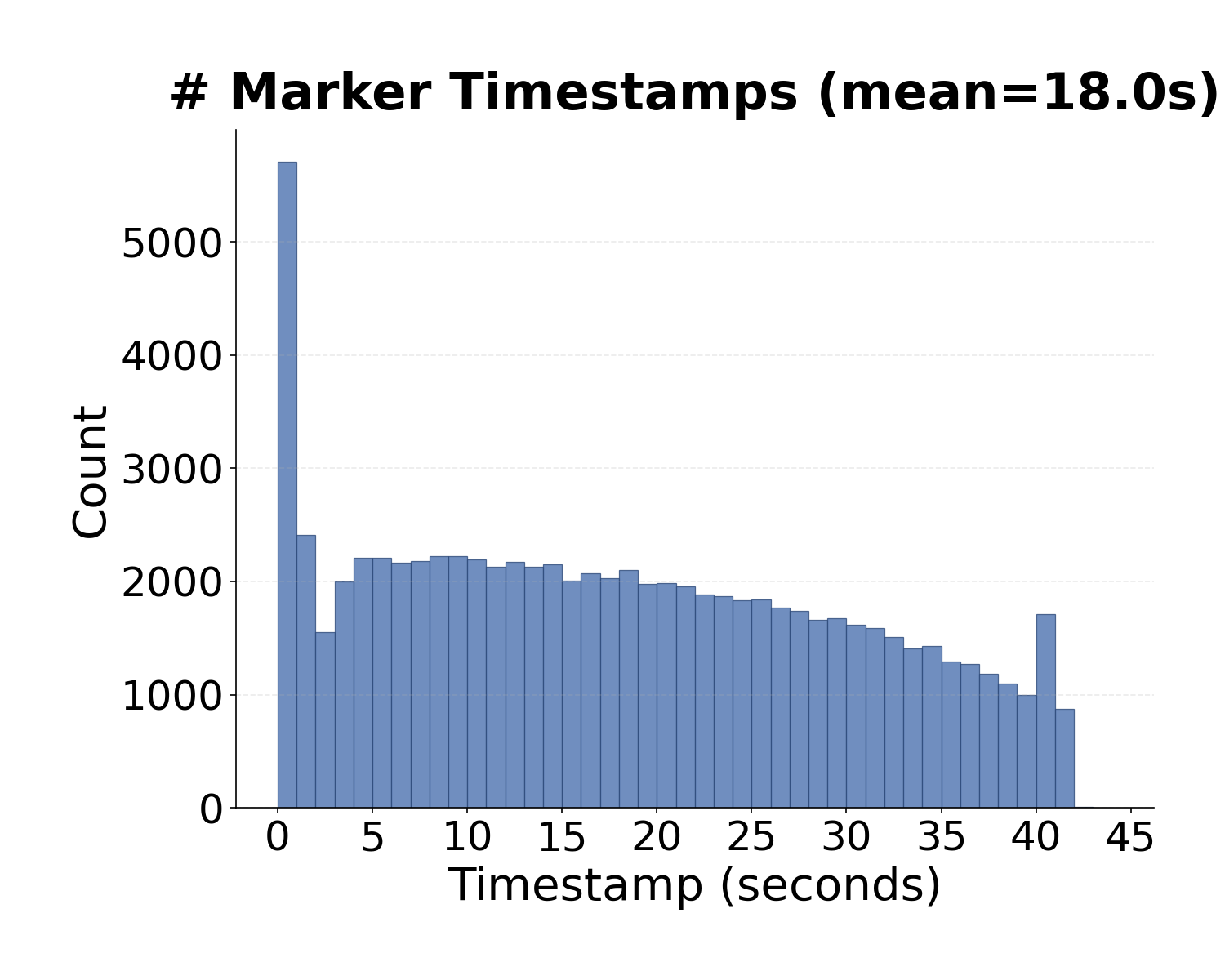}
        \caption{Marker Timestamps}
        \label{fig:tm-timestamps}
    \end{subfigure}
    \hfill
    \begin{subfigure}[t]{0.32\textwidth}
        \centering
        \includegraphics[width=\textwidth]{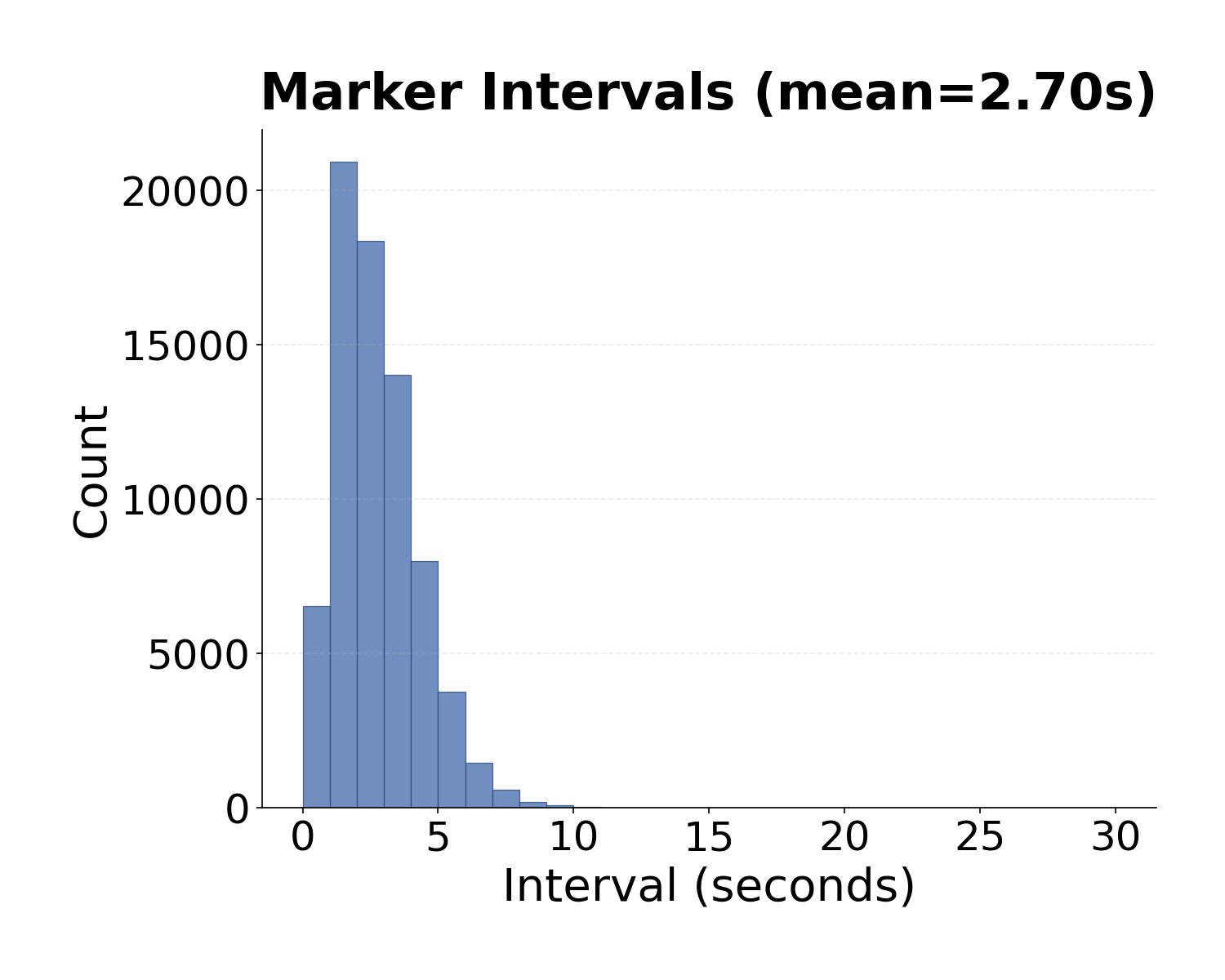}
        \caption{Inter-Marker Intervals}
        \label{fig:tm-intervals}
    \end{subfigure}
    \caption{Distribution of Spoken Time Markers in the First stage training data.}
    \label{fig:time-marker-stats}
\end{figure*}

%% file: Tables/main_results_text.tex
\begin{table*}[h]
\centering
\caption{
\textbf{Text-query} evaluation on LIFEBench and UROBench-text 
under Short (10--30\,s) and Long (30--60\,s) settings. Results 
are reported as MAE (seconds) / MAPE (\%), with the rightmost 
column showing GPT-score averaged across both benchmarks. Lower 
is better for MAE/MAPE, and higher is better for GPT-score. 
Model categories are indicated by color: 
\legenddot{cascadedcolor}~Cascaded,
\legenddot{commercialcolor}~Commercial,
\legenddot{opensourcecolor}~Open-sourced, and
\legenddot{proposedcolor}~Proposed. 
\textbf{Bold} marks the best result per column.
}
\label{tab:text_q_results}
\renewcommand{\arraystretch}{1.2}

\resizebox{\textwidth}{!}{%
\begin{tabular}{lcccccc}
\toprule

& \multicolumn{2}{c}{Short (10s--30s)}
& \multicolumn{2}{c}{Long (30s--1min)}
& \multicolumn{2}{c}{Overall} \\
\cmidrule(lr){2-3} \cmidrule(lr){4-5} \cmidrule(lr){6-7}

\textbf{Model}
& \textbf{LIFEBench}
& \textbf{UROBench-text}
& \textbf{LIFEBench}
& \textbf{UROBench-text}
& \textbf{MAE~/~MAPE}
& \textbf{GPT-score}\\
\midrule

\cascadedbox{Cascade (GPT)}
& 4.83 / 23.5\%
& 5.92 / 28.8\%
& 8.35 / 18.4\%
& 19.44 / 41.7\%
& 9.64 / 28.1\% 
& \textbf{3.58} \\

\cascadedbox{Cascade (Qwen)}
& 7.98 / 40.4\%
& 10.38 / 51.2\%
& 19.16 / 42.9\%
& 29.72 / 64.2\%
& 16.81 / 49.7\% 
& 2.86 \\

\commercialbox{GPT-audio}
& 14.33 / 73.6\% 
& 18.75 / 97.1\%
& 11.07 / 26.3\%
& 15.94 / 38.0\%
& 15.02 / 58.8\% 
& 2.78 \\

\opensourcebox{LFM Audio}
& 18.84 / 101.1\%
& 12.43 / 60.4\%
& 35.74 / 79.1\%
& 39.21 / 85.4\%
& 26.56 / 81.5\% 
& 1.95 \\

\opensourcebox{Mimo-Audio}
& 12.66 / 62.0\%
& 9.40 / 46.8\%
& 31.75 / 69.7\%
& 29.59 / 63.8\%
& 20.85 / 60.6\% 
& 1.39 \\

\midrule
\opensourcebox{Qwen2.5-Omni-7B}
& 13.74 / 79.1\%
& 7.73 / 44.9\%
& 13.86 / 30.8\%
& 19.18 / 40.6\%
& 13.63 / 48.9\% 
& 2.67 \\

\proposedbox{\textbf{TiCo (Proposed)}}
& \textbf{4.46 / 22.7\%}
& \textbf{4.08 / 21.0\%}
& \textbf{6.62 / 14.8\%}
& \textbf{6.25 / 13.5\%}
& \textbf{5.35 / 18.0\%} 
& 2.76 \\

\bottomrule
\end{tabular}%
}

\end{table*}













%% file: Figures/fig_text.tex

\begin{figure*}[h]
    \centering
    \includegraphics[width=\textwidth]{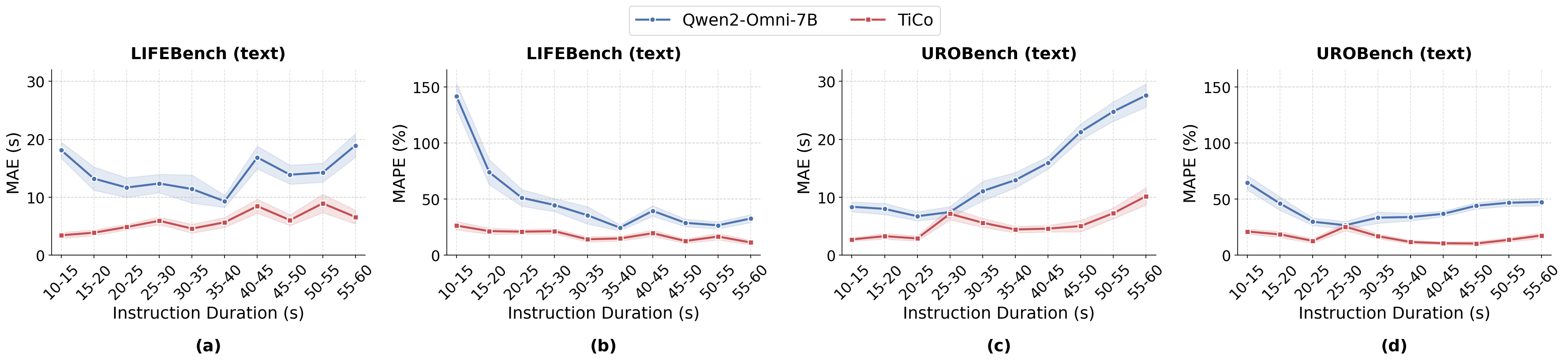}
    \caption{Text benchmarks: duration error of Qwen2.5-Omni-7B vs.\ TiCo measured against instructed duration. From left to right: LIFEBench MAE~(s), LIFEBench MAPE~(\%), UROBench MAE~(s), UROBench MAPE~(\%). Shaded regions indicate $\pm$1 SEM.}
    \label{fig:text-vs-target}
\end{figure*}

\begin{figure*}[h]
    \centering
    \includegraphics[width=\textwidth]{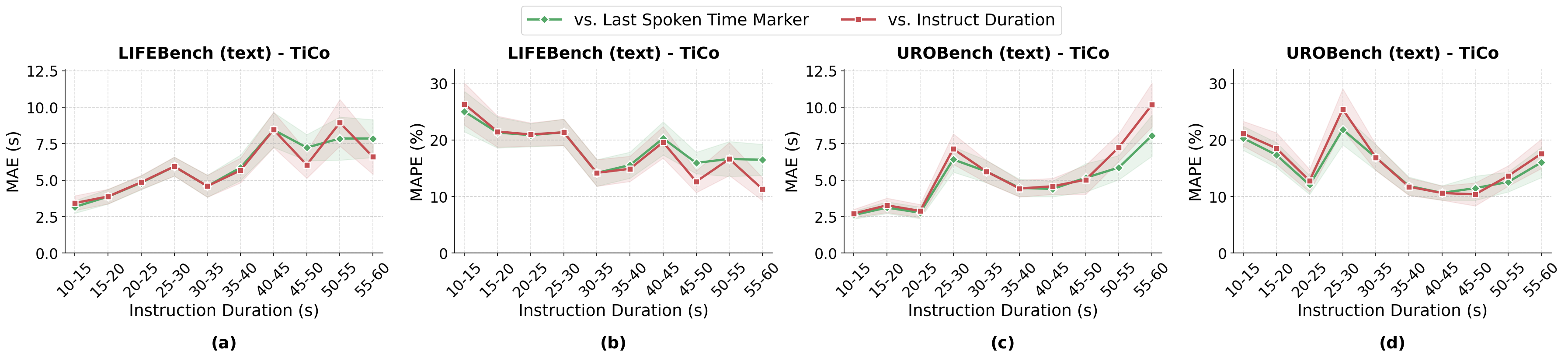}
    \caption{Text benchmarks (TiCo): duration error measured against instructed duration vs.\ last time marker. From left to right: LIFEBench MAE~(s), LIFEBench MAPE~(\%), UROBench MAE~(s), UROBench MAPE~(\%). Shaded regions indicate $\pm$1 SEM.}
    \label{fig:text-vs-token}
\end{figure*}

%% file: Tables/per_task_mae.tex
\begin{table*}[ht]
\centering
\caption{Per-task \textbf{MAE} (seconds) on TiCo-Bench. Lower is better.
Model categories indicated by color:
\legenddot{cascadedcolor}~Cascaded,
\legenddot{commercialcolor}~Commercial,
\legenddot{opensourcecolor}~Open-sourced,
and \legenddot{proposedcolor}~Proposed.
\textbf{Bold} marks the best result in each column.
$^{*}$For Kimi Audio on the SUM subset, audio clips longer than 120\,s exceed the model's effective context window and are excluded from evaluation.}
\label{tab:mae-detail}
\renewcommand{\arraystretch}{1.15}
\resizebox{\textwidth}{!}{%
\begin{tabular}{>{\raggedright\arraybackslash}m{3.4cm} cccc cccc c}
\toprule
\emph{TiCo-Bench (Speech)}
& \multicolumn{4}{c}{Short (10s--30s)}
& \multicolumn{4}{c}{Long (30s--60s)}
& Overall \\
\cmidrule(lr){2-5} \cmidrule(lr){6-9} \cmidrule(lr){10-10}
\textbf{Model}
& \textbf{QA} & \textbf{REA} & \textbf{CRE} & \textbf{SUM}
& \textbf{QA} & \textbf{REA} & \textbf{CRE} & \textbf{SUM}
& \textbf{MAE $\downarrow$} \\
\midrule

\cascadedbox{Cascade (Qwen)}
& 7.59 & 10.45 & 7.32 & 10.35
& 21.48 & 30.27 & 17.92 & 11.47
& 14.61 \\

\cascadedbox{Cascade (GPT)}
& 3.96 & 5.70 & 3.80 & \textbf{3.26}
& 11.33 & 18.47 & 12.19 & 7.91
& 8.46 \\

\commercialbox{GPT-audio}
& 16.74 & 20.72 & 23.71 & 11.99
& 19.90 & 20.63 & 21.60 & 17.80
& 19.36 \\

\opensourcebox{Kimi Audio}
& 24.03 & 29.26 & 28.01 & 10.49$^{*}$
& 26.77 & 31.79 & 26.62 & 21.86$^{*}$
& 26.50 \\

\opensourcebox{LFM Audio}
& 15.61 & 12.37 & 12.00 & 7.85
& 22.41 & 34.36 & 24.78 & 22.66
& 19.88 \\

\opensourcebox{MiMo-Audio}
& 13.77 & 12.50 & 21.60 & 11.51
& 20.42 & 17.53 & 17.96 & 15.29
& 16.37 \\

\opensourcebox{Qwen3-Omni-30B}
& 9.10 & 8.87 & 12.05 & 5.26
& 13.98 & 17.72 & 11.58 & 16.30
& 12.02 \\
\midrule

\opensourcebox{Qwen2.5-Omni-7B}
& 8.32 & 7.91 & 9.60 & 13.05
& 15.58 & 19.96 & 14.18 & 11.48
& 12.57 \\

\proposedbox{\textbf{TiCo (Proposed)}}
& \textbf{3.00} & \textbf{3.37} & \textbf{3.42} & 8.10
& \textbf{5.30} & \textbf{6.38} & \textbf{5.86} & \textbf{5.94}
& \textbf{4.70} \\

\bottomrule
\end{tabular}
}
\end{table*}

%% file: Tables/per_task_gptscore.tex
\begin{table*}[t]
\centering
\caption{Per-task \textbf{GPT-score} (1--5 scale) on TiCo-Bench. Higher is better.
Model categories indicated by color:
\legenddot{cascadedcolor}~Cascaded,
\legenddot{commercialcolor}~Commercial,
\legenddot{opensourcecolor}~Open-sourced,
and \legenddot{proposedcolor}~Proposed.
\textbf{Bold} marks the best result in each column.
$^{*}$For Kimi Audio on the SUM subset, audio clips longer than 120\,s exceed the model's effective context window and are excluded from evaluation.}
\label{tab:gpt-detail}
\renewcommand{\arraystretch}{1.15}
\resizebox{\textwidth}{!}{%
\begin{tabular}{>{\raggedright\arraybackslash}m{3.4cm} cccc cccc c}
\toprule
\emph{TiCo-Bench (Speech)}
& \multicolumn{4}{c}{Short (10s--30s)}
& \multicolumn{4}{c}{Long (30s--60s)}
& Overall \\
\cmidrule(lr){2-5} \cmidrule(lr){6-9} \cmidrule(lr){10-10}
\textbf{Model}
& \textbf{QA} & \textbf{REA} & \textbf{CRE} & \textbf{SUM}
& \textbf{QA} & \textbf{REA} & \textbf{CRE} & \textbf{SUM}
& \textbf{GPT-score $\uparrow$} \\
\midrule

\cascadedbox{Cascade (Qwen)}
& 3.34 & 3.27 & 3.56 & 2.64
& 3.21 & 3.29 & 3.17 & 2.48
& 3.18 \\

\cascadedbox{Cascade (GPT)}
& \textbf{4.07} & \textbf{4.11} & \textbf{4.13} & \textbf{2.88}
& \textbf{4.11} & \textbf{4.19} & \textbf{3.99} & \textbf{2.86}
& \textbf{3.92} \\

\commercialbox{GPT-audio}
& 3.74 & 3.74 & 3.38 & 2.78
& 3.56 & 3.33 & 3.06 & 2.50
& 3.45 \\

\opensourcebox{Kimi Audio}
& 1.93 & 1.56 & 2.03 & 1.10
& 1.72 & 1.46 & 1.92 & 1.10
& 1.69 \\

\opensourcebox{LFM Audio}
& 3.09 & 2.22 & 2.91 & 1.67
& 2.85 & 2.18 & 2.73 & 1.49
& 2.58 \\

\opensourcebox{MiMo-Audio}
& 2.48 & 3.19 & 3.20 & 2.07
& 3.24 & 3.22 & 3.15 & 1.96
& 2.91 \\

\opensourcebox{Qwen3-Omni-30B}
& 3.89 & 3.68 & 3.63 & 2.20
& 3.79 & 3.57 & 3.75 & 1.87
& 3.58 \\
\midrule

\opensourcebox{Qwen2.5-Omni-7B}
& 3.55 & 3.48 & 3.59 & 2.38
& 3.29 & 3.37 & 3.07 & 2.37
& 3.31 \\

\proposedbox{\textbf{TiCo (Proposed)}}
& 3.31 & 3.50 & 3.40 & 2.34
& 3.39 & 3.67 & 3.31 & 2.36
& 3.32 \\

\bottomrule
\end{tabular}
}
\end{table*}

%% file: Tables/per_task_utmos.tex
\begin{table*}[t]
\centering
\caption{Per-task \textbf{UTMOS} (1--5 scale) on TiCo-Bench. Higher is better.
The first row reports the ground-truth query speech as a reference and is excluded from the bold comparison.
Model categories indicated by color:
\legenddot{cascadedcolor}~Cascaded,
\legenddot{commercialcolor}~Commercial,
\legenddot{opensourcecolor}~Open-sourced,
and \legenddot{proposedcolor}~Proposed.
\textbf{Bold} marks the best result in each column.
$^{*}$For Kimi Audio on the SUM subset, audio clips longer than 120\,s exceed the model's effective context window and are excluded from evaluation.}
\label{tab:utmos-detail}
\renewcommand{\arraystretch}{1.15}
\resizebox{\textwidth}{!}{%
\begin{tabular}{>{\raggedright\arraybackslash}m{3.4cm} cccc cccc c}
\toprule
\emph{TiCo-Bench}
& \multicolumn{4}{c}{Short (10s--30s)}
& \multicolumn{4}{c}{Long (30s--60s)}
& Overall \\
\cmidrule(lr){2-5} \cmidrule(lr){6-9} \cmidrule(lr){10-10}
\textbf{Model}
& \textbf{QA} & \textbf{REA} & \textbf{CRE} & \textbf{SUM}
& \textbf{QA} & \textbf{REA} & \textbf{CRE} & \textbf{SUM}
& \textbf{UTMOS $\uparrow$} \\
\midrule

GT (Speech Query)
& 4.28 & 3.79 & 4.16 & 3.90
& 4.28 & 3.79 & 4.16 & 3.90
& 4.08 \\
\midrule

\cascadedbox{Cascade (Qwen)}
& 3.03 & 3.17 & 3.03 & 2.88
& 2.92 & 3.07 & 2.86 & 2.75
& 3.00 \\

\cascadedbox{Cascade (GPT)}
& 3.00 & 3.07 & 3.03 & 2.93
& 2.78 & 2.89 & 2.80 & 2.69
& 2.91 \\

\commercialbox{GPT-audio}
& 3.89 & 3.82 & 3.73 & 3.86
& 3.65 & 3.54 & 3.55 & 3.50
& 3.72 \\

\opensourcebox{Kimi Audio}
& 2.47 & 2.56 & 2.30 & 2.59$^{*}$
& 2.31 & 2.49 & 2.25 & 2.48$^{*}$
& 2.43 \\

\opensourcebox{LFM Audio}
& 3.96 & 4.14 & 3.98 & 4.03
& 3.96 & 4.11 & 3.90 & 3.99
& 4.01 \\

\opensourcebox{MiMo-Audio}
& 3.06 & 3.47 & 3.49 & 3.55
& 3.26 & 3.47 & 3.25 & 3.41
& 3.31 \\

\opensourcebox{Qwen3-Omni-30B}
& \textbf{4.44} & \textbf{4.45} & \textbf{4.42} & \textbf{4.45}
& \textbf{4.44} & \textbf{4.45} & \textbf{4.43} & \textbf{4.44}
& \textbf{4.44} \\
\midrule

\opensourcebox{Qwen2.5-Omni-7B}
& 4.11 & 4.12 & 4.07 & 4.07
& 4.09 & 4.10 & 4.03 & 4.03
& 4.09 \\

\proposedbox{\textbf{TiCo (Proposed)}}
& 4.10 & 4.11 & 4.07 & 4.12
& 3.98 & 3.99 & 3.97 & 4.03
& 4.04 \\

\bottomrule
\end{tabular}
}
\end{table*}

%% file: Tables/timemarker_mae.tex
\begin{table}[h]
\centering
\caption{Local alignment quality of Spoken Time Markers in TiCo.
Each cell reports the mean absolute error in seconds between the predicted
Spoken Time Marker and the corresponding ASR-aligned word timestamp.}
\label{tab:stm-alignment}
\renewcommand{\arraystretch}{1.1}
\begin{tabular}{lcc}
\toprule
Task & Short (10s--30s) & Long (30s--60s) \\
\midrule
QA            & 1.90 & 3.09 \\
Reasoning     & 2.08 & 3.32 \\
Creative      & 2.05 & 3.60 \\
Summarization & 7.45 & 3.62 \\
\midrule
Overall       & 2.65 & 3.21 \\
\bottomrule
\end{tabular}
\end{table}

%% file: Tables/qualitative.tex
\label{app:duration_examples}

\subsection{How deep is the ocean? (Speech query)}

\begin{responseboxblue}{How deep is the ocean?, target duration = 15 s}
\textbf{Prompt:} \emph{How deep is the ocean? Please respond with exactly 15 seconds.}

\medskip
\textbf{Speech response:}

{\ttfamily\small
Well,~\ts{0.9} you know,~\ts{1.6} the ocean is really deep in some places.
\ts{3.8} The Mariana Trench is the deepest part,~\ts{7.2} it's about 36 kilometers deep.
\ts{9.4} But overall,~\ts{11.0} the average depth of the world's oceans is around 3.7 kilometers.~\ts{15.0}
}

\medskip
\textbf{Target duration:} \emph{15.0 seconds}
\\
\textbf{Actual duration:} \emph{15.2 seconds}
\end{responseboxblue}

\begin{responseboxblue}{How deep is the ocean?, target duration = 25 s}
\textbf{Prompt:} \emph{How deep is the ocean? Please respond with exactly 25 seconds.}

\medskip
\textbf{Speech response:}

{\ttfamily\small
Well,~\ts{1.0} you know,~\ts{1.9} the ocean is really deep in some places.
\ts{4.3} The Mariana Trench is the deepest part,~\ts{7.6} it's about 36,000 feet or so deep.
\ts{10.8} But overall,~\ts{12.6} the average depth of the world's oceans is around 13,000 feet.
\ts{16.4} It's amazing how vast and deep they are,~\ts{19.2} right?~\ts{20.0}
If you want to know more about the ocean's depths or anything else related,~\ts{25.0}
}

\medskip
\textbf{Target duration:} \emph{25.0 seconds}
\\
\textbf{Actual duration:} \emph{25.0 seconds}
\end{responseboxblue}

\begin{responseboxblue}{How deep is the ocean?, target duration = 40 s}
\textbf{Prompt:} \emph{How deep is the ocean? Please respond with exactly 40 seconds.}

\medskip
\textbf{Speech response:}

{\ttfamily\small
Well,~\ts{1.5} you know,~\ts{2.3} the ocean is really deep in some places.
\ts{4.6} The Mariana Trench is one of the deepest parts.
\ts{7.9} It's about 36 kilometers deep.
\ts{10.8} But overall,~\ts{12.5} if we're talking about an average depth for the whole ocean,~\ts{15.5} it's around 3.7 kilometers.
\ts{18.0} However,~\ts{19.4} different areas can have very different depths.
\ts{22.3} Some shallow bays might be just a few meters deep while others in the middle of the ocean could be much deeper than that.
\ts{29.0} So,~\ts{30.0} it's not really a fixed number like 3.7 kilometers everywhere.
\ts{33.0} If you want to know more about why the ocean has such varying depths or anything else related,~\ts{39.0} feel free to ask.~\ts{40.0}
}

\medskip
\textbf{Target duration:} \emph{40.0 seconds}
\\
\textbf{Actual duration:} \emph{41.6 seconds}
\end{responseboxblue}

\subsection{Discuss an event from history (Speech query)}

\begin{responseboxblue}{History event, target duration = 10 s}
\textbf{Prompt:} \emph{Discuss an event from history. Please respond with exactly 10 seconds.}

\medskip
\textbf{Speech response:}

{\ttfamily\small
Well,~\ts{1.5} let's talk about the fall of the Berlin Wall in 1989.\\
\ts{4.2} It was a huge event that symbolized the end of the Cold War.\\
\ts{7.3} It brought East and West Germany together again.~\ts{10.0}
}

\medskip
\textbf{Target duration:} \emph{10.0 seconds}
\\
\textbf{Actual duration:} \emph{12.0 seconds}
\end{responseboxblue}

\begin{responseboxblue}{History event, target duration = 20 s}
\textbf{Prompt:} \emph{Discuss an event from history. Please respond with exactly 20 seconds.}

\textbf{Speech response:}

{\ttfamily\small
Well,~\ts{1.5} let's talk about the signing of the Magna Carta in 1215.
\ts{4.8} It was really important because it limited the power of the king and protected the rights of the nobles.
\ts{9.6} It also had an impact on later democratic ideas.
\ts{13.7} You know,~\ts{15.0} it set a precedent for the rule of law.
\ts{18.0} So what do you think about this event?~\ts{20.0}
}

\medskip
\textbf{Target duration:} \emph{20.0 seconds}
\\
\textbf{Actual duration:} \emph{19.2 seconds}
\end{responseboxblue}

\begin{responseboxblue}{History event, target duration = 30 s}
\textbf{Prompt:} \emph{Please respond with exactly 30 seconds.}

\textbf{Speech response:}

{\ttfamily\small
Well,~\ts{1.5} let's talk about the signing of the Magna Carta in 1215.
\ts{4.8} It was really important because it limited the power of the king to some extent.
\ts{9.6} You know,~\ts{10.7} before that kings could pretty much do whatever they wanted with their subjects' rights.
\ts{14.9} But this document said things like no one can be imprisoned without proper trial.
\ts{19.6} It set a precedent for later legal systems all over the world.
\ts{24.0} And it kind of started the idea of due process.
\ts{28.0} So what do you think about this event?~\ts{30.0}
}

\medskip
\textbf{Target duration:} \emph{30.0 seconds}
\\
\textbf{Actual duration:} \emph{31.1 seconds}
\end{responseboxblue}





\subsection{What is quantum mechanics? (Text query)}

\begin{responseboxgreen}{Quantum mechanics, target duration = 25 s}
\textbf{Prompt:} \emph{What is quantum mechanics? Please respond with exactly 25 seconds.}

\medskip
\textbf{Speech response:}

{\ttfamily\small
Well,~\ts{1.0} quantum mechanics is this really strange science that deals with really small things like atoms and subatomic particles.
\ts{7.0} It has some wacky ideas,~\ts{9.0} you know,~\ts{10.0} like particles can be in multiple places at once or they can be entangled so that if one changes state the other does too instantaneously no matter how far apart.
\ts{23.0} It's super important for understanding how the world works on a very fundamental level.\ts{25.0}
}
\\\\
\textbf{Target duration:} \emph{25.0 seconds }
\\
\textbf{Actual duration:} \emph{29.1 seconds}
\end{responseboxgreen}

\begin{responseboxgreen}{Quantum mechanics, target duration = 50 s}
\textbf{Prompt:} \emph{What is quantum mechanics? Please respond with exactly 50 seconds.}

\medskip
\textbf{Speech response:}

{\ttfamily\small
Well,~\ts{1.9} quantum mechanics is really something special.
\ts{4.3} It's a branch of physics that deals with the behavior of really small things like atoms and subatomic particles.
\ts{12.6} You know,~\ts{14.0} it has some really strange ideas.
\ts{16.0} For example,~\ts{17.8} particles can be in multiple places at once until you measure them.
\ts{21.8} And there are these things called wave-particle duality where particles act both like waves and particles.
\ts{29.2} Also,~\ts{30.6} the Heisenberg uncertainty principle says that we can't precisely know certain pairs of properties at the same time.
\ts{37.8} It's all very counterintuitive compared to what we see in our everyday world.
\ts{43.0} If you want to learn more about specific aspects or have any other questions,~\ts{48.0}
}
\\\\
\textbf{Target duration:} \emph{50.0 seconds }
\\
\textbf{Actual duration:} \emph{44.0 seconds}
\end{responseboxgreen}

\subsection{Why is Mars considered a candidate for human colonization? (Text query)}

\begin{responseboxgreen}{Mars colonization, target duration = 10 s}
\textbf{Prompt:} \emph{Why is Mars considered a candidate for human colonization? Please respond with exactly 10 seconds.}

\medskip
\textbf{Speech response:}

{\ttfamily\small
Well,~\ts{1.5} Mars has resources like water ice,~\ts{3.8} which could be used for life support.
\ts{5.4} It also has an atmosphere that might be modified to make it more habitable.
\ts{9.2} So yeah,~\ts{10.0} it's a good candidate.
}

\medskip
\textbf{Target duration:} \emph{10.0 seconds}
\\
\textbf{Actual duration:} \emph{12.0 seconds}
\end{responseboxgreen}

\begin{responseboxgreen}{Mars colonization, target duration = 30 s}
\textbf{Prompt:} \emph{Why is Mars considered a candidate for human colonization? Please respond with exactly 30 seconds.}

\medskip
\textbf{Speech response:}

{\ttfamily\small
Well,~\ts{1.5} Mars has some things going for it.
\ts{4.2} It's closer to Earth than other potential places like Jupiter's moons.
\ts{8.6} There might be water ice on its surface,~\ts{11.7} which could be used for life support systems.
\ts{14.9} Also,~\ts{16.3} its gravity is lower compared to Earth's,~\ts{19.0} making it easier to build structures there.
\ts{21.7} And it has an atmosphere that can potentially be modified over time.
\ts{25.5} So all these factors make Mars a good candidate for human colonization.~\ts{30.0}
}

\medskip
\textbf{Target duration:} \emph{30.0 seconds}
\\
\textbf{Actual duration:} \emph{27.9 seconds}
\end{responseboxgreen}

\begin{responseboxgreen}{Mars colonization, target duration = 60 s}
\textbf{Prompt:} \emph{Why is Mars considered a candidate for human colonization? Please respond with exactly 60 seconds.}

\medskip
\textbf{Speech response:}

{\ttfamily\small
Well,~\ts{1.5} Mars is considered a candidate for human colonization for several reasons.
\ts{4.9} First off,~\ts{6.2} it has some resources like water ice at the poles which could be used to make oxygen and hydrogen for fuel.
\ts{13.7} Also,~\ts{15.0} there might be minerals on Mars that we can mine in the future.
\ts{18.5} The gravity on Mars is about one-third of Earth's,~\ts{22.2} so it would be easier for humans to move around compared to being in zero gravity.
\ts{27.0} And scientists think that over time,~\ts{30.0} we could potentially modify the Martian environment to be more habitable.
\ts{34.0} For example,~\ts{35.6} building habitats that can protect against radiation and extreme temperatures.
\ts{40.0} There is also the possibility of finding signs of past life on Mars,~\ts{44.0} which makes it even more interesting.
\ts{46.0} If you want to know more about this or have other questions related to space exploration,~\ts{50.0} feel free to ask.~\ts{51.0}
}

\medskip
\textbf{Target duration:} \emph{60.0 seconds}
\\
\textbf{Actual duration:} \emph{57.4 seconds}
\end{responseboxgreen}

%% file: Figures/pattern.drawio.tex
\begin{figure*}[h!]
    \centering
    \includegraphics[width=\textwidth]{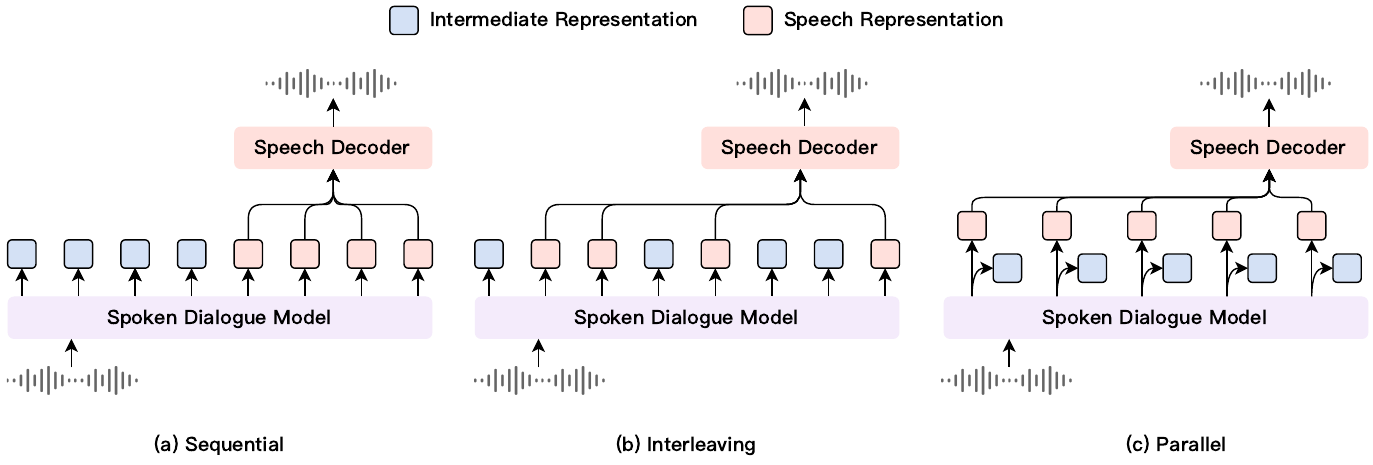}
    \caption{Illustration of different generation patterns in spoken dialogue models (SDMs): (a) Sequential, (b) Interleaved, and (c) Parallel. }
    \label{fig:Pattern-SDM}
\end{figure*}

%% file: Tables/survey.tex
\begin{table*}[h!]
\centering
\small
\setlength{\tabcolsep}{3pt}
\caption{Spoken dialogue models (SDMs) with speech input and speech output, ordered by their first public release time. 
\textbf{Date}: First released date.
\textbf{IR}: Intermediate representation used in the SDM. 
\textbf{Speech Rrep.}: Speech representation (prediction target) used by the model. 
\textbf{Pattern}: The pattern of how the intermediate and speech representations are generated.}
\label{tab:sdm_ir_survey}
\resizebox{\columnwidth}{!}{
\begin{tabular}{lccccp{5.5cm}}
\toprule
\textbf{Model} & \textbf{Date} & \textbf{IR} & \textbf{Speech Rrep.} & \textbf{Pattern} & \textbf{Notes} \\
\midrule

dGSLM~\cite{nguyen2023generative} & 2022-03 & - & Phonetic token & Direct & ``Dual-tower'' architecture for dual channel full-duplex modeling. Direct modeling of two-channel phonetic tokens. \\
SpeechGPT~\cite{zhang2023speechgpt} & 2023-05 & Text (+R.) & Phonetic token & Sequential & Using ``Chain-of-Modality''. Expands LLM vocabulary with phonetic tokens. \\
SPIRIT-LM~\cite{nguyen2025spirit} & 2024-02 & Text (+S.) & Phonetic token & Interleaved & Interleaves text and speech in one stream; expressive version adds pitch and style tokens.\\
Moshi~\cite{defossez2024moshi} & 2024-07 & Text & Acoustic token & Parallel & Dual-channel full-duplex model. Parallel decoding text and acoustic tokens with delay pattern. \\
LLaMA-Omni~\cite{fangllama} & 2024-09 & Text & Phonetic token & Parallel & CTC speech decoder maps LLM response states to phonetic tokens for streaming speech synthesis. \\
SyncLLM~\cite{veluri2024beyond} & 2024-09 & - & Phonetic token & Direct & Interleaving user speech and model speech for full-duplex dialogue. \\

Mini-Omni2~\cite{xie2024mini} & 2024-10 & Text & Acoustic token & Parallel & Parallel decoding with delay pattern. \\
Freeze-Omni~\cite{wang2025freeze} & 2024-11 & Text & Acoustic token & Sequential & TDM-based full-duplex interaction; speech decoder conditioned on text tokens and LLM hidden states. \\

GLM-4-Voice~\cite{zeng2024glm} & 2024-12 & Text & Phonetic token & Interleaved & Single speech codebook paired with flow matching speech decoder.  \\
SLAM-Omni~\cite{chen2025slam} & 2024-12 & Text & Phonetic token & Parallel & ``Semantic group modeling'' enables generating multiple phonetic tokens per text token. \\

VITA-1.5~\cite{fu2025vita} & 2025-01 & Text & Acoustic token & Sequential &  NAR + AR speech decoder taking LLM embedding as input.\\

Baichuan-Audio~\cite{li2025baichuan} & 2025-02 & Text & Acoustic token & Interleaved & Text-guided speech generation with an independent audio head. \\

Qwen2.5-Omni~\cite{xu2025qwen25} & 2025-03 & Text & Acoustic token & Sequential & Thinker-Talker architecture. Supports visual
modality.\\

Kimi-Audio~\cite{ding2025kimi} & 2025-04 & Text & Phonetic token & Parallel & Shared LLM with text head and audio head.\\

LLaMA-Omni 2~\cite{fang2025llama} & 2025-05 & Text & Phonetic token & Parallel & Gate fusion of LLM hidden states and text tokens for improved speech quality.\\

Step-Audio 2~\cite{wu2025step} & 2025-07 & Text (+R., Tool)& Phonetic token & Interleaved & Using ``multi-modal RAG''  to support grounded response and timbre/style control.\\

STITCH~\cite{chiang2026stitch} & 2025-07 & Text (+R.) & Phonetic token & Interleaved & Backbone: GLM-4-Voice. Various interleaving text, reasoning, speech patterns are discussed in the paper. \\

Qwen3-Omni~\cite{xu2025qwen3} & 2025-09 & Text (+R.)& Acoustic token & Sequential & Thinker-Talker architecture. Supports explicit thinking mode and compatible with tool calling. Supports visual modality. \\

Moshi-CoT~\cite{shih2025can} & 2025-10 & Text (+R.) & Acoustic token & Parallel & CoT-tuned Moshi performs text reasoning in the ``text monologue'' stream to enable ``thinking while listening'' paradigm.\\

SCoT~\cite{arora2025chain} & 2025-10 & Text & Acoustic token & Interleaved & CoT framework for SDMs. Blockwise streaming full-duplex model. \\

Streaming RAG~\cite{arora2025stream} & 2025-10 & Text (+Tool) & Acoustic token & Sequential & Enables the SDM to trigger tool queries in parallel with the user's speech. \\

LFM2-Audio~\cite{amini2025lfm2} & 2025-11 & Text & Acoustic token & Interleaved & Supports both interleaved and sequential patterns, adapting to different tasks.\\

Mimo-Audio~\cite{zhang2025mimo} & 2025-12 & Text (+R.) & Acoustic token & Interleaved & Interleaving text tokens and ``audio patches'', which includes a delay pattern. \\

PersonaPlex~\cite{roy2026personaplex} & 2026-01 & Text & Acoustic token & Parallel & Followed the Moshi architecture. Dual-channel full-duplex model.\\

MiniCPM-o 4.5~\cite{minicpm_o} & 2026-02 & Text (+R.)& Acoustic token & Interleaved & TDM full-duplex model. Supports visual modality.\\

\bottomrule
\end{tabular}
}
\end{table*}